\documentclass{article}

    \PassOptionsToPackage{numbers, compress}{natbib}
\usepackage[preprint]{neurips_2024}

\usepackage[utf8]{inputenc} % allow utf-8 input
\usepackage[T1]{fontenc}    % use 8-bit T1 fonts

\usepackage[colorlinks=true, linkcolor=NavyBlue,   citecolor=NavyBlue, urlcolor=NavyBlue]{hyperref}       % hyperlinks
\usepackage{url}            % simple URL typesetting
\usepackage[dvipsnames]{xcolor}         % colors
\usepackage{booktabs}       % professional-quality tables
\usepackage{amsfonts}       % blackboard math symbols
\usepackage{nicefrac}       % compact symbols for 1/2, etc.
\usepackage{microtype}      % microtypography

\newcommand{\proposed}{TFMoE}
\usepackage{tablefootnote}
\usepackage{multirow}

\usepackage[normalem]{ulem}
\useunder{\uline}{\ul}{}
\usepackage{enumitem}
\usepackage{array}
\usepackage{caption}
\usepackage{multirow}
\usepackage{graphicx}
\usepackage{wrapfig}
\usepackage{booktabs}
\usepackage{adjustbox}
\usepackage{lineno}
\usepackage{algorithm} 
\usepackage{algpseudocode}
\usepackage{placeins}
\usepackage{pifont}
\newcommand{\cmark}{\ding{51}}%
\newcommand{\xmark}{\ding{55}}%

\usepackage{wrapfig}
\usepackage{xcolor}
\usepackage{amsmath,amsfonts,bm}

\def\eqref#1{equation~\ref{#1}}
\def\1{\bm{1}}

\DeclareMathAlphabet{\mathsfit}{\encodingdefault}{\sfdefault}{m}{sl}
\SetMathAlphabet{\mathsfit}{bold}{\encodingdefault}{\sfdefault}{bx}{n}

\newcommand{\TableMain}{
\begin{table*}[t]
\caption{Performance of traffic flow forecasting on PEMSD3-Stream dataset. The model names are appended with their corresponding pre-existing sensor access ratios, e.g., \proposed~ (1\%) indicates the \proposed~ model with a 1\% access ratio. Importantly, $\gamma$ represents the ratio of pre-existing sensors accessed due to the subgraph structure. Please note that, in line with standard practice in traffic forecasting research, we use the exact prediction values corresponding to their exact time frames: 15-minute, 30-minute, and 60-minute, for their respective evaluation metrics\protect\footnotemark.}
\vspace{-1ex}
\fontsize{7pt}{8}\selectfont
\begin{adjustbox}{center}
\begin{tabular}{@{}cc|c@{\hspace{5.5pt}}c@{\hspace{5.5pt}}c|c@{\hspace{5.5pt}}c@{\hspace{5.5pt}}c|c@{\hspace{5.5pt}}c@{\hspace{5.5pt}}c|cll@{}}

\toprule

\multicolumn{2}{c|}{PEMSD3-Stream}                                         & \multicolumn{3}{c|}{15 min}                      & \multicolumn{3}{c|}{30 min}                      & \multicolumn{3}{c|}{60 min}                      & \multicolumn{3}{c}{\multirow{2}{*}{\begin{tabular}[c]{@{}c@{}}Training Time\\ (sec)\end{tabular}}} \\ \cmidrule(r){1-11}
\multicolumn{2}{c|}{Model}                                          & MAE            & RMSE           & MAPE           & MAE            & RMSE           & MAPE           & MAE            & RMSE           & MAPE           & \multicolumn{3}{c}{}                                                                               \\ \midrule
\multicolumn{1}{c|}{\multirow{4}{*}{Retrained}} & GRU               & 13.51          & 21.83          & 18.19          & 15.80          & 25.89          & 21.05          & 23.47          & 35.53          & 28.50          & \multicolumn{3}{c}{1824}                                                                        \\
\multicolumn{1}{c|}{}                           & DCRNN             & 12.39          & 19.09          & 17.42          & 13.88          & 21.50          & 19.51          & 17.37          & 26.79          & 24.70          & \multicolumn{3}{c}{50K+}                                                                           \\
\multicolumn{1}{c|}{}                           & STSGCN            & 12.97          & 21.07          & 17.92          & 13.81          & 22.54          & 18.90          & 15.89          & 25.92          & 21.93          & \multicolumn{3}{c}{50K+}                                                                           \\
\multicolumn{1}{c|}{}                           & Retrained-\proposed  & 12.25          & 20.11          & 16.17          & 13.61          & 22.42          & 17.56          & 16.59          & 27.17          & 21.38          & \multicolumn{3}{c}{940}                                                                         \\ \midrule
\multicolumn{1}{c|}{\multirow{4}{*}{Online}}    & Static-\proposed     & 12.95          & 21.18          & 18.97          & 14.51          & 23.90          & 19.62          & 18.07          & 29.87          & 24.92          & \multicolumn{3}{c}{194}                                                                         \\
\multicolumn{1}{c|}{}                           & Expansible-\proposed & 13.55          & 21.92          & 19.53          & 15.30          & 24.94          & 21.36          & 19.51          & 31.72          & 27.32          & \multicolumn{3}{c}{228}                                                                         \\
\multicolumn{1}{c|}{}                           & TrafficStream (10+$\gamma$\%)     & 13.84          & 22.52          & 28.99          & 15.96          & 26.13          & 22.83          & 21.22          & 34.22          & 32.07          & \multicolumn{3}{c}{239}                                                                         \\
\cmidrule(l){2-14} 
\multicolumn{1}{c|}{}                           & \proposed~(1\%)           & \textbf{12.48} & \textbf{20.48} & \textbf{16.72} & \textbf{13.93} & \textbf{23.00} & \textbf{18.13} & \textbf{17.20} & \textbf{28.36} & \textbf{22.17} & \multicolumn{3}{c}{244}                                                                \\ \bottomrule
\end{tabular}

\end{adjustbox}
\label{tab:table_main}
\end{table*}
}

\title{Continual Traffic Forecasting via Mixture of Experts}

\author{
\textbf{Sanghyun Lee} \hspace{1cm} \textbf{Chanyoung Park}\\
Korea Advanced Institute of Science and Technology (KAIST)\\
\texttt{\{ragdoll1762, cy.park\}@kaist.ac.kr}\\
}

\begin{document}

\maketitle

\begin{abstract}
The real-world traffic networks undergo expansion through the installation of new sensors, implying that the traffic patterns continually evolve over time.
Incrementally training a model on the newly added sensors would make the model forget the past knowledge, i.e., catastrophic forgetting, while retraining the model on the entire network to capture these changes is highly inefficient. 
To address these challenges, we propose a novel Traffic Forecasting Mixture of Experts (\proposed) for traffic forecasting under evolving networks.
The main idea is to segment the traffic flow into multiple homogeneous groups, and assign an expert model responsible for a specific group. 
This allows each expert model to concentrate on learning and adapting to a specific set of patterns, while minimizing interference between the experts during training, thereby preventing the dilution or replacement of prior knowledge, which is a major cause of catastrophic forgetting.
Through extensive experiments on a real-world long-term streaming network dataset, PEMSD3-Stream, we demonstrate the effectiveness and efficiency of~\proposed. Our results showcase superior performance and resilience in the face of catastrophic forgetting, underscoring the effectiveness of our approach in dealing with continual learning for traffic flow forecasting in long-term streaming networks.
\end{abstract}

\section{Introduction}
\label{sec:intro}
Recently, numerous studies have been proposed with the goal of enhancing the accuracy of traffic forecasting.
However, while various models have been proposed to tackle this problem~\citep{dcrnn,zgcnnet,stgcn,wavenet,stgode,st-gart,gman,astgcn,gts,stemgnn,agcrn,stfgnn,stag-gcn,dstagnn,stsgcn}, most of them focus on improving accuracy in \textit{static} traffic networks.

In this work, we focus on the real-world traffic forecasting scenarios, where the traffic networks undergo expansion through the installation of new sensors in the surrounding areas (i.e., \textit{evolving} traffic network).
While these newly added sensors may exhibit traffic patterns similar to pre-existing ones, they also introduce previously unobserved patterns.
Moreover, even pre-existing sensors may display new patterns over long-term periods (e.g., several years) due to various factors such as urban development, infrastructural projects, or population migration. 
Consequently, if a model that is trained on the past traffic network is further incrementally trained on the newly added sensors in the expanded network, the model would forget the past knowledge, resulting in a severe performance degradation on the pre-existing sensors in the past network, which is called catastrophic forgetting. A straightforward solution would be to re-train the model on the entire dataset containing not only the newly added sensors but also the pre-existing sensors.
However, the process of retraining the model is computationally demanding and time-consuming, highlighting the necessity for a more suitable and efficient learning methodology.

TrafficStream~\citep{trafficstream} is a pioneering work that focuses on traffic forecasting under evolving traffic networks. Its main idea is to adopt continual learning strategies to continuously learn and adapt from ongoing data streams, integrating new information while preserving the past knowledge (i.e., avoid catastrophic forgetting).
However, despite its effectiveness, the dynamic nature of traffic networks continues to pose significant challenges, highlighting the necessity for a more adaptive approach to accommodate these diverse and evolving traffic patterns. Specifically, TrafficStream adopts a ``one-model-fits-all'' approach, which uses a single model to capture all the evolving traffic patterns, being susceptible to catastrophic forgetting.

\begin{wrapfigure}{r}{0.4\textwidth} 
  \centering
  \includegraphics[width=0.4\textwidth]{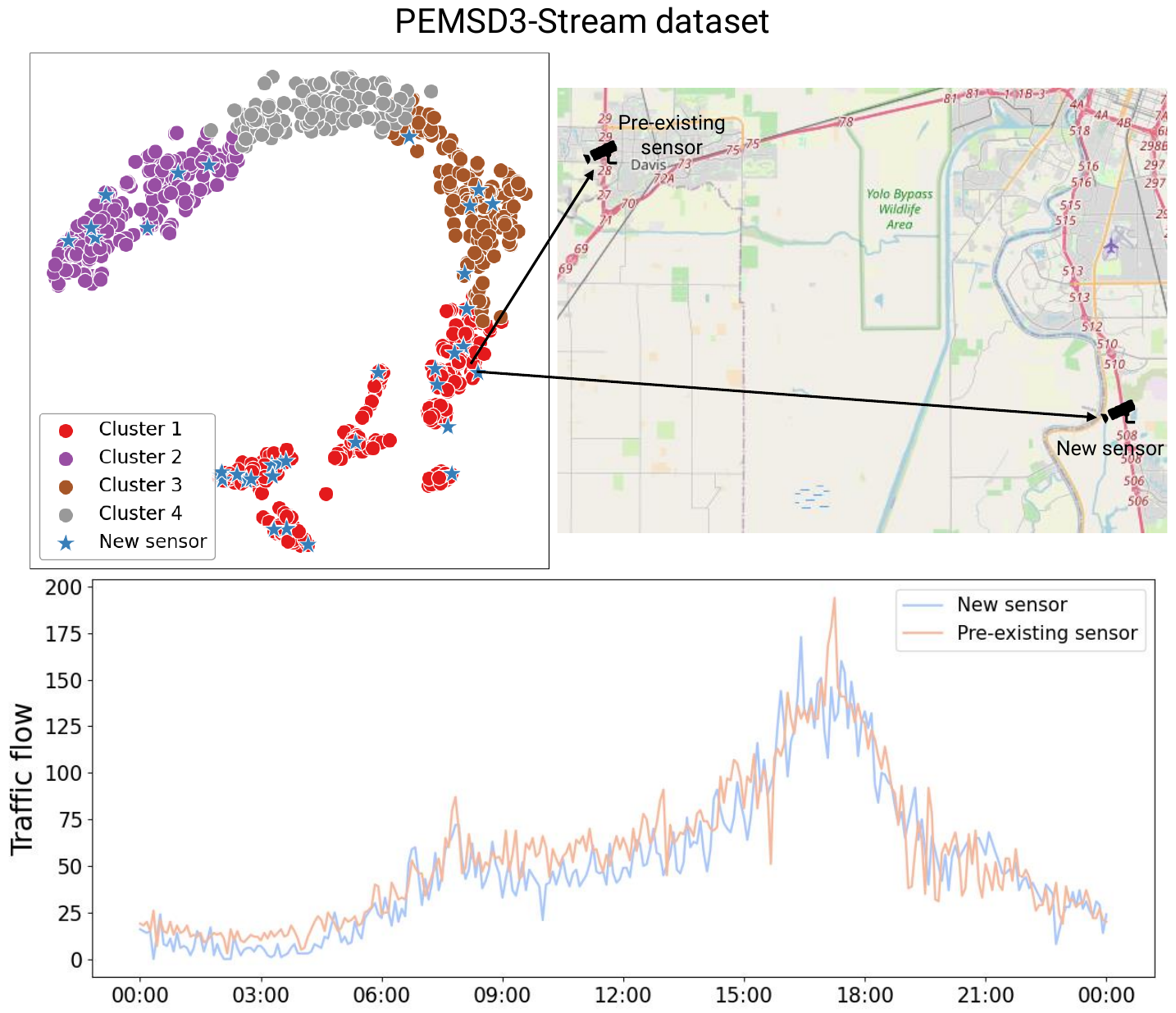} 
  \caption{The top-left plot depicts the t-SNE visualization of one week's traffic patterns gathered from each sensor in the traffic network, with newly added sensors of the next year marked by stars. The top-right image shows the geographical location of a new sensor and its closest counterpart in the latent space. The bottom plot indicates the notable similarity between the traffic patterns obtained from these two sensors. }
  \label{fig_latent}
    \vspace{-1ex}
\end{wrapfigure}

To this end, we propose Traffic Forecasting Mixture of Experts (\proposed) for traffic forecasting under evolving networks. 
The main idea of~\proposed~is to segment the traffic data into multiple homogeneous groups, regardless of the geographical distances on the map, and assign an expert traffic forecasting model to each group.
By doing so, we allow each model to be responsible for predicting the traffic flow of its assigned group, minimizing interference with each other during training, which in turn alleviates catastrophic forgetting. 
This is possible because each model can concentrate on learning and adapting to a specific set of patterns, thereby preventing the dilution or replacement of prior knowledge, which is a major cause of catastrophic forgetting. Figure~\ref{fig_latent} demonstrates the motivation of our work. We observe that the sensors can be segmented into multiple homogeneous groups based on their traffic patterns, and moreover the newly added sensors tend to belong to one of the existing clusters. That is, even if the traffic network is expanded, the newly added sensors exhibit similar traffic patterns as those of pre-existing sensors. This implies that having an expert solely dedicated to each homogeneous group would be more effective than the ``one-model-fits-all'' approach in terms of alleviating catastrophic forgetting. This is because each expert only needs to concentrate on learning and adapting to a specific set of patterns, while the ``one-model-fits-all'' approach needs to adapt to the global dynamics even though local changes mainly occur.

Each expert model in~\proposed~contains two components: 1) a reconstructor that utilizes a Variational Autoencoder (VAE) structure to reconstruct the traffic flow, and 2) a predictor that makes future predictions based on the past traffic flow.
We first cluster the traffic flow based on the representation extracted by a pre-trained feature extractor, and train an expert model (i.e., a reconstructor and a predictor) on each cluster. 
Then, when a new traffic flow is introduced, we assign it to an expert model whose reconstruction loss is the smallest.
Finally, we make the final prediction by combining the individual predictions from each expert model using a reconstruction-based gating mechanism.

At the core of our approach are three pivotal strategies.  1) `\textit{Reconstructor Based Knowledge Consolidation Loss},' inspired by the learning without forgetting (LwF)~\citep{cl_reg_lwf} approach, ensures that the model learns new traffic patterns while also preserving knowledge from previous tasks based on the concept of the localized group within the VAE. 2) `\textit{Forgetting Resilient Sampling}' addresses catastrophic forgetting by generating synthetic data through decoders of earlier-trained reconstructors. Within the VAE framework, this synthetic data, being both similar in nature but rich in diversity, is used alongside current task nodes for training. While the generated data might not inherently represent geographical graph structures, our graph learning technique ensures seamless integration.
3) `\textit{Reconstruction Based Replay}' employs a reconstructor to detect sensors that exhibit patterns not familiar to any expert. These nodes, determined by their reconstruction probability spanning all experts, are merged with the current task nodes, creating a dataset that captures patterns previously elusive to our expert models.

Through extensive experiments on a real-world long-term streaming network dataset, PEMSD3-Stream, we highlight the advantages of~\proposed. Our results demonstrate that~\proposed~outperforms existing models, demonstrating significantly better performance and resilience against catastrophic forgetting. These findings validate the superiority of our proposed approach in addressing the challenges associated with continual learning in long-term streaming network scenarios, providing a robust and effective solution for traffic forecasting.

\section{RELATED WORK}
\vspace{-1ex}
\noindent\textbf{Traffic Flow Forecasting. }
In traffic forecasting, traditional methods like ARIMA and SVR are popular but often miss intricate spatio-temporal patterns in road networks due to their reliance on historical data. Deep learning, specifically the combination of RNNs and CNNs as in \citep{rnncnn,rnncnn2,rnncnn3}, has emerged to address these complexities. However, CNNs, being optimized for grid data, are not ideal for road networks. 
Thus, the focus has shifted to GCNs \citep{gcn,gcn2,gcn3,gat} for capturing spatial relationships \citep{dcrnn,zgcnnet,stgcn,wavenet,stgode,st-gart,gman,astgcn,gts,stemgnn,agcrn,stfgnn,stag-gcn,dstagnn,stsgcn}. 
Additionally, there exist methods based on the experts ~\cite{expert_traffic_1,expert_traffic_2}, and methods based on clustering~\cite{cluster1, cluster2}. Despite progress, most research remains on static traffic networks, overlooking the evolving nature of real-world traffic.

\noindent\textbf{Continual Learning. }
Continual learning, or lifelong learning, targets systems that adapt to changing environments and accumulate knowledge, aiming primarily to prevent catastrophic forgetting—losing old knowledge while learning new. This domain has three main strategies: regularization-based, rehearsal-based, and architecture-based. Regularization methods~\citep{cl_reg_ewc, cl_reg_si} add terms to the loss function, limiting model parameter changes. Rehearsal methods~\citep{cl_mem_er,cl_mem_icarl,cl_generative_1,cl_generative_2} utilize replay buffers or generative models to retain past data or tasks. Architecture strategies~\citep{cl_arc_pnn,cl_arc_packnet} dynamically alter the model structure for new tasks.
While continual learning has been recently explored in the graph domain~\citep{cl_graph_1,cl_graph_3,cl_graph_4}, a majority of studies focus on classification rather than regression, making direct application to streaming traffic networks challenging~\citep{trafficstream,cl_traffic_1,pecpm}. 

\vspace{-1ex}
\section{Problem Definition}

In the context of long-term streaming traffic networks, we define $\tau \in ({1,2, \ldots, \mathcal{T}})$ as an extended time interval, or a `Task,' where the traffic network remains unchanged. These dynamic networks are sequenced as $G=\left(G^1, G^2, \ldots, G^\mathcal{T}\right)$, with each $G^\tau$ evolving from its predecessor $G^{\tau-1}$ via $G^\tau=G^{\tau-1}+\Delta G^\tau$.
For a task $\tau$, its road network is defined as $G^{\tau}=(V^{\tau}, A^{\tau})$, where $V^{\tau}$ is the set with $N^{\tau}$ traffic sensors and $A^{\tau} \in \mathbb{R}^{N^{\tau} \times N^{\tau}}$ is its adjacency matrix. Network variations are captured by $\Delta G^\tau=(\Delta V^{\tau}, \Delta A^{\tau})$, with $\Delta V^{\tau}$ denoting new nodes.
While the graph structure $G$ can rely on 
geographical-distance-based graphs,
% geographical-based graphs, 
learning (or inferring) the structure from data is better suited for continual learning, as detailed in Section \ref{sec:training_predictor}.

For a specific task denoted as $\tau$, the observed traffic flow data across the node set $N^{\tau}$ during the time span $\left(t-T^{\prime}+1\right):t$ is represented by $\mathbf{X}^{\tau}_t = \{x^{\tau}_{1,t}, x^{\tau}_{2,t}, ..., x^{\tau}_{N^{\tau},t}\}\in \mathbb{R}^{N^{\tau} \times T^{\prime}}$.
In this context, each element $x^{\tau}_{i,t}\in \mathbb{R}^{T^{\prime}}$ represents the data for the $i^{th}$ node, spanning the preceding $T^{\prime}$ time steps starting from time $t$.
Likewise, data covering the subsequent time interval $(t+1):(t+T)$ is represented as $\mathbf{Y}^{\tau}_t = \{y^{\tau}_{1,t}, y^{\tau}_{2,t}, ..., y^{\tau}_{N^{\tau},t}\}\in \mathbb{R}^{N^{\tau} \times T}$, where each $y^{\tau}_{i,t}\in \mathbb{R}^{T}$ represents the upcoming $T$ time steps from $t+1$ for the $i^{th}$ node.
Additionally, we introduce $\mathbf{X}^{\tau}_{w} = \{x^{\tau}_{1,w}, x^{\tau}_{2,w}, ..., x^{\tau}_{N^{\tau},w}\}\in \mathbb{R}^{N^{\tau} \times week}$, where \textit{week} refers to the total time steps obtained by segmenting an entire week into time intervals. Consequently, each $x^{\tau}_{i,w}\in \mathbb{R}^{week}$ denotes the initial one-week traffic data for the $i^{th}$ node. We select the first full week of data from Monday to Sunday in the training dataset.

Our primary objective is to develop a probabilistic regression model, parameterized by $\theta$, that can predict $\mathbf{Y}^{\tau}_t$ using both $\mathbf{X}^{\tau}_t$ and $\mathbf{X}^{\tau}_{w}$. This can be formally represented as
$
p\left(\mathcal{D}^{\tau} \mid \theta\right)=\prod_{t} \prod_{i = 1}^{N^{\tau}} p\left(y^{\tau}_{i,t} \mid x^{\tau}_{i,t},x^{\tau}_{i,w}; \theta\right),
$
where $\mathcal{D}^\tau=\left\{\left(\mathbf{Y}^{\tau}_t, \mathbf{X}^{\tau}_t,\mathbf{X}^{\tau}_{w}\right)\right\}_t$ denotes the dataset corresponding to task $\tau$.
Based on the above notations, we now describe our main approach—the Mixture of Experts ~\citep{moe} (MoE) framework. 
For a system with $K$ experts, the probability distribution $p\left(y^{\tau}_{t} \mid {x}^{\tau}_{t},{x}^{\tau}_{w}; \theta\right)$, based on a gating mechanism, can be described as follows:
\vspace{-2pt}
\begin{equation}
p\left(y^{\tau}_{t} \mid {x}^{\tau}_{t},{x}^{\tau}_{w}; \theta\right)=\sum_{k=1}^K
\underbrace{p\left({y}^{\tau}_{t} \mid {x}^{\tau}_{t},\eta=k; \theta\right)}
_{\begin{array}{c}
\text { predictor }
\end{array}} 
\underbrace{p\left( \eta=k \mid {x}^{\tau}_{w};\theta\right)}
_{\begin{array}{c}
\text { gating term }
\end{array}} 
, 
\label{eq:prob2}
\end{equation}
where $\eta$ denotes the expert indicator, and $p\left( \eta=k \mid {x}^{\tau}_{w};\theta\right)$ denotes the probability assigned by the gating mechanism to expert $k$ for the given input $x^{\tau}_w$ . 
The central challenge emerges when modeling the gating term $p\left( \eta=k \mid {x}^{\tau}_{w};\theta\right)$.
Rather than employing a single classifier model for modeling the gating term, our approach, inspired by~\citep{moe_generative}, adopt the generative modeling: 
$
p\left(y^{\tau}_{t} \mid {x}^{\tau}_{t},{x}^{\tau}_{w}; \theta\right)=
\sum_{k=1}^K
p\left(y^{\tau}_{t} \mid {x}^{\tau}_{t} ; \psi_k\right) \frac{p\left({x}^{\tau}_{w}  ; \phi_k\right) p(\eta=k)}{\sum_{k^{\prime}} p\left({x}^{\tau}_{w} ;  \phi_{k^{\prime}}\right) p\left(\eta=k^{\prime}\right)}.
$
Here, the function $p\left({x}^{\tau}_{w}; \phi_k\right)$ represents the generative model, which is a reconstructor in our context. 
For clarity, we split the parameter $\theta$ into expert-specific parameters, i.e., $\theta=\cup_{k=1}^K {\theta}_{k} \text {, where } \theta_{k}=\left\{\psi_{k}, \phi_k\right\}$.
Additionally, we assume that for $p\left(\eta=k\right)$, the probability remains uniform. 
In the setting of long-term streaming traffic networks, the objective is to learn the sequence $\left(\theta^1, \theta^2, \ldots, \theta^\mathcal{T}\right)$.
Given the context of our continual learning scenario, it is important to note that during the learning of task $\tau$, we use parameters initialized with $ \theta^{\tau -1}$ and conduct the learning process based on $\Delta N^\tau=\left|\Delta V^\tau\right|$. 
\vspace{-2ex}
\section{Proposed Method:~\proposed}
\vspace{-1ex}
\subsection{Pre-training Stage}
\label{sec:pretraining}
\vspace{-1ex}
\subsubsection{Reconstruction-Based Clustering}
\label{sec:reconstruction_based_clustering}
\vspace{-1ex}
\begin{wrapfigure}{R}{0.45\textwidth} 
\vspace{-5ex}
  \centering
  \includegraphics[width=0.45\textwidth]{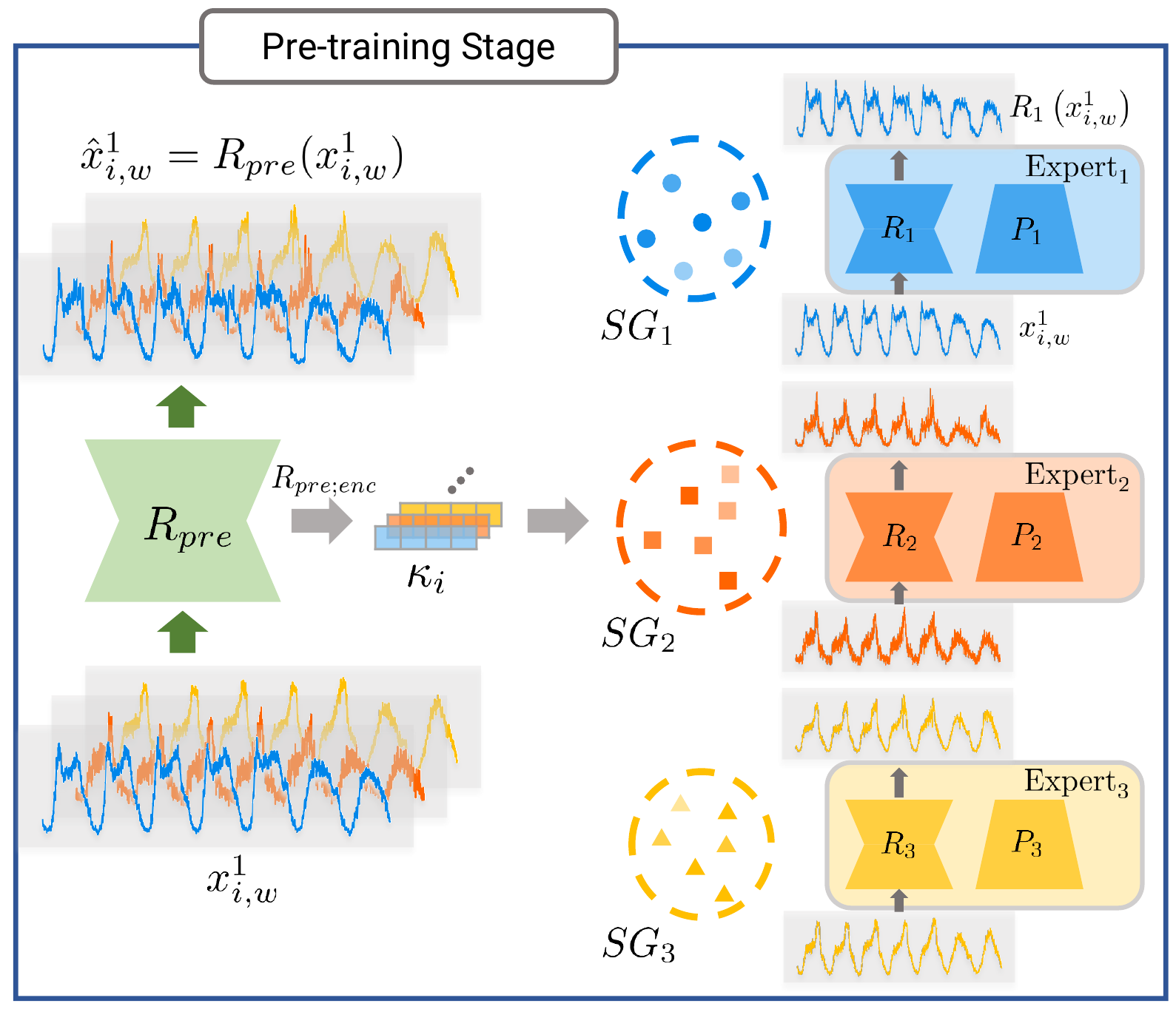} 
  \vspace{-2ex}
  \caption{Architecture of Pre-training Stage.}
  \label{fig_main1}
  \vspace{-4.5ex}
\end{wrapfigure} 
In the initial phase of our training pipeline, we aim to group the $N^1 =| V^1 | $ sensors from the first task (i.e., $\tau = 1$) into $K$ homogeneous clusters. We apply Deep Embedded Clustering~\citep{cluster_dec}, which enables dual learning of features and cluster assignments via deep networks. First, we pre-train an autoencoder feature extractor using a week's traffic data from each sensor. The reconstructed output for sensor $i$ is $\hat{x}^1_{i, w}=R_{{pre}}(x^1_{i, w})$, where $R_{{pre}}$ is the autoencoder and $x^1_{i, w}$ is the week-long data from sensor $i$. For optimization, we employ the MAE loss $\mathcal{L}_{recon}=\frac{1}{N^1} \sum_{i=1}^{N^1} \left\|{x}^{1}_{i, w}- \hat{x}^1_{i, w}\right\|_1.$

After training the feature extractor $R_{{pre}}$, we encode the week-long traffic data from sensor $i$, $x^{1}_{i, w}$, to a latent representation $\kappa_i=R_{pre;{enc}}(x^{1}_{i, w})$, where $\kappa_i \in \mathbb{R}^{d_{\mathcal{Z}}}$. $R_{pre;{enc}}$ is the encoder part of autoencoder $R_{{pre}}$. Using these representations, we perform k-means clustering to get $K$ cluster centroids $\left[\mu_1 ; \ldots ; \mu_K\right] \in \mathbb{R}^{K \times d_{\mathcal{Z}}}$, which serve as initial learnable parameters. We measure the soft cluster assignment probability between $\kappa_i$ and centroid $\mu_k$ using the Student's t-distribution~\citep{cluster_student}: 
$
q_{i k}=\frac{\left(1+\left\|\kappa_i-\mu_k\right\|^2\right)^{-{1}}}{\sum_{k^{\prime}}\left(1+\left\|\kappa_i-\mu_{k^{\prime}}\right\|^2\right)^{-{1}}} ,
$
where $q_{i k}$ denotes the probability of assigning sensor $i$ to cluster $k$.
We further refine clustering via an auxiliary target distribution $p_i$ as follows~\citep{cluster_dec}: 
$
p_{i k}=\frac{q_{i k}^2 / \sum_{i^{\prime}} q_{i^{\prime} k}}{\sum_{k^{\prime}}\left(q_{i k^{\prime}}^2 / \sum_{i^{\prime}} q_{i^{\prime} k^{\prime}}\right)}.
$
The distribution $p_{i k}$ is strategically designed to augment the homogeneity within clusters, while giving precedence to data points associated with high confidence levels.
For the purpose of achieving high-confidence assignments, we define the KL divergence loss between $q_i$ and $p_i$ as $\mathcal{L}_{cluster}=D_{K L}(P \| Q)=\sum_i \sum_k p_{i k} \log \frac{p_{i k}}{q_{i k}}$.
Optimizing both the reconstruction loss $\mathcal{L}_{recon}$ and the clustering loss $\mathcal{L}_{cluster}$ aids in extracting meaningful patterns from data and clustering sensors with similar characteristics. The overall loss is defined as: \(\mathcal{L}_{p} = \mathcal{L}_{recon} + \alpha\mathcal{L}_{cluster}\), where \(\alpha\) is the weight of the clustering loss.

\vspace{-1ex}
\subsubsection{Constructing Experts}
\vspace{-1ex}
Utilizing the cluster assignment probability $q_i$, we established a hard assignment $c_i = {\operatorname{argmax}}_{k}\left(q_{i k}\right)$ for each sensor $i$. 
Then, the group of sensors assigned to the $k$-th cluster is defined as $S G_k=\left\{i \mid c_i=k\right\}$, and 
sensors that belong to the same cluster shares homogeneous semantics. Accordingly, we assign an expert, i.e., $ \text{Expert}_k = ( R_k, P_k ) $ comprising of a reconstructor $R_k$ and a predictor $P_k$, to each cluster $k$.

\vspace{-1ex}
\subsubsection{Training Reconstructor of Expert}
\vspace{-1ex}
\label{sec:training_reconstructor}
Under the continual learning framework, we aim to train the reconstructor with two objectives:
\vspace{-1ex}
\begin{enumerate}[leftmargin=0.5cm]
\item{\textbf{Sensor-Expert Matching.}}
In the ever-evolving traffic network landscape, the integration of a new sensor mandates the identification of an expert that is semantically compatible. To this end, we train the reconstructor $R_k$ to proficiently reconstruct the representations of its designated sensor group, denoted as $SG_k$. This strategic training ensures that, upon the introduction of new sensors, we can seamlessly identify the most appropriate expert that aligns with its semantic content.

\item{\textbf{Generate Synthetic Samples.}} One of the primary concerns as we transition to subsequent tasks is to minimize catastrophic forgetting. Instead of directly storing the current task data in memory, we utilize a latent variable $z_k$ for each $\text{Expert}_k$ to generate data via $p\left(x_w \mid z_k; \phi_k\right)$. 
However, as the posterior distribution $p\left(z_k \mid x_w; \phi_k\right)$ is intractable, an approximation through the variational posterior $q\left(z_k \mid x_w; \xi_k\right)$, which is parameterized by $\xi_k$, becomes essential.
\end{enumerate}
\vspace{-0.5ex}

To address the aforementioned objectives, we model each expert's reconstructor as a Variational Autoencoder (VAE) \citep{vae}. For the $k$-th expert, the marginal likelihood for all data in $SG_k$ is given by:
$
\log p\left(x_{(S G_k[1],w)}, \ldots, x_{(S G_k \left[|SG_k|\right],w)};\phi_k\right) 
=\sum_{i \in S G_k} \log p\left(x_{i,w};\phi_k\right).
$
For each data point $i$, it unfolds as:
\vspace{-0.1ex}
\begin{equation}
\begin{aligned}
\log p & \left(x_{i,w};\phi_k\right) = D_{K L}\left(q\left(z_k \mid x_{i,w};{\xi_k}\right) \| p\left(z_k \mid x_{i,w};{\phi_k}\right)\right)\\
& \quad  + \mathbb{E}_{q\left(z_k \mid x_{i,w};{\xi_k}\right)}\left[-\log q\left(z_k \mid x_{i,w};{\xi_k}\right)+\log p(x_{i,w}, z_k;\phi_k)\right].
\end{aligned}
\label{eq:prob3}
\end{equation}

The Evidence Lower Bound (ELBO) for expert $k$ is then given by: 
$
\mathcal{L}_{ELBO}^k\left(\phi_k,\xi_k\right)=
\sum_{i \in S G_k}
\mathbb{E}_{q\left(z_k \mid x_{i,w};{\xi_k}\right)}\left[\log \frac{p\left(x_{i,w} \mid z_k;\phi_k\right)p(z_k;{\phi_k})}{q\left(z_k \mid x_{i,w};{\xi_k}\right)}\right].
$
Training follows a similar approach to conventional VAEs, where the objective is to maximize $\mathcal{L}_{ELBO}^k$ for each cluster. Consequently, the overall objective for all clusters becomes:
$
\mathcal{L}_{ELBO}^{SG} = \sum_{k =1}^K \mathcal{L}_{ELBO}^k\left(\phi_k,\xi_k\right). 
$
We set the prior using learnable parameters $\mu_k$ and $\Sigma_k$ as $p(z_k;{\phi_k})=\mathcal{N}\left(z_k \mid \mu_k, \Sigma_k\right)$ to enhance the expressiveness of the latent space for the data assigned to each expert, where  $\Sigma_k$ is a diagonal matrix. The method for sampling is described in detail in Section~\ref{sec:forgetting_resilient_sampling}.

\begin{wrapfigure}{R}{0.6\textwidth} 
\vspace{-3ex}
  \centering
  \includegraphics[width=0.6\textwidth]{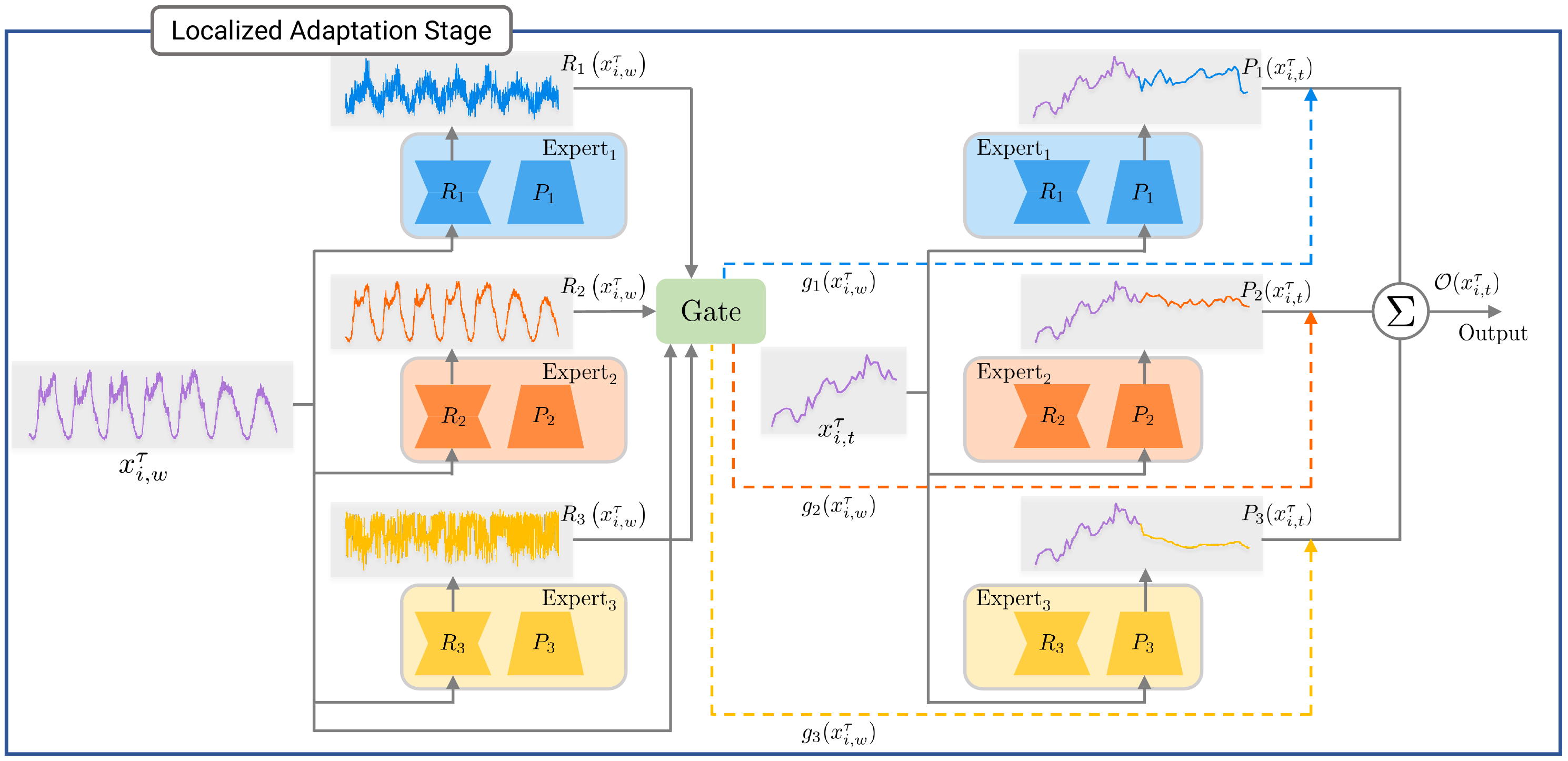} 
  \vspace{-1ex}
  \caption{Architecture of Localized Adaptation Stage.}
  \vspace{-2.5ex}
  \label{fig_main2}
\end{wrapfigure} 

\vspace{-1ex}
\subsection{Localized Adaptation Stage}
\label{sec:localized}
\vspace{-1ex}
\subsubsection{Training Predictor of Expert}
\label{sec:training_predictor}
\vspace{-1ex}
We use the Mixture of Experts~\citep{moe} framework to train each expert's predictor. This predictor takes the past $T^{\prime}$ time steps of data $x_{i, t}^\tau$ from the $i^{th}$ sensor to forecast the next $T$ time steps $y_{i, t}^\tau$. We employ Graph Neural Network (GNN) layers to capture spatial sensor dependencies and 1D-Convolutional layers for the temporal dynamics of traffic. 
However, we argue that using predefined geographical-distance-based graphs for GNNs presents the following two issues:
\vspace{-1ex}
\begin{enumerate}[leftmargin=0.5cm]
    \item {\textbf{Handling Newly Added Nodes.}} When learning from newly added nodes within predefined graph structures rooted in actual geographical distances, Graph Neural Networks (GNNs) require the formation of \textit{`subgraphs'} centered around these newly added nodes. Consequently, these newly added nodes $\Delta V^\tau$ inevitably establish many connections with pre-existing nodes $V^{(\tau-1)}$ from previous tasks. 
    For this reason, when learning the current task using a GNN, it needs to access many pre-existing nodes. Although accessing as much data of pre-existing nodes as possible is indeed beneficial, it violates the goal of continual learning, whose main goal is to achieve optimal performance with minimal access to previous tasks.

    \item {\textbf{Lack of Graph Structure in Sampled Data.}} Recall that to consolidate prior knowledge, we will sample data from a VAE decoder, $R_k$, previously trained on an earlier task (i.e., Forgetting-Resilient Sampling, which will be described in Section \ref{sec:forgetting_resilient_sampling}). However, as they are synthetically generated, they inherently lack the graph structural information, implying that the generated nodes would be isolated in a predefined geographical-distance-based graph. As a result, GNNs are hindered from maximizing their inherent strength of information propagation across nodes.
\end{enumerate}
\vspace{-1ex}

\noindent\textbf{Solution: Graph Structure Learning. }
We address these issues by adopting the graph structure learning mechanism~\citep{gsl}, leveraging the Gumbel softmax trick~\citep{gumbel,gumbel2}, instead of using predefined geographical-distance-based graphs. For a given node $i$, its hidden embedding can be represented as: $e_i = L_e(x_{i,t}^{\tau})$, where 
$L_e(\cdot)$ is a linear transformation. Subsequently, the weight connecting nodes $i$ and $j$ is modeled as: $w_{ij} = L_w([e_i ; e_j]) -\ln(-\ln(U))$, where $L_w(\cdot)$ is also a linear transformation. In this equation, $U$ is drawn from a uniform distribution, $U \sim \text{Uniform}(0, 1)$, which serves to introduce Gumbel noise into the model. In essence, the adjacency matrix of the graph that links nodes $i$ and $j$ can be represented as $A_{ij} = \frac{\exp(w_{ij})}{\sum_{j^\prime} \exp(w_{ij^\prime})}$.
The learned adjacency matrix inherently exhibits non-symmetry. To address this characteristic effectively, we adopt the Diffusion Convolution Layer~\citep{dcrnn}; a Graph Neural Network (GNN) layer designed to model spatial dependencies via a diffusion process. Given an input graph signal $\mathcal{X} \in \mathbb{R}^{N \times D}$ and an adjacency matrix $A_{ij}$, the diffusion convolution operation can be defined as: 
$
\mathcal{X}_{:, p} \star_{A} f_{\boldsymbol{\zeta}}=\sum_{m=1}^{M}\left(\boldsymbol{\zeta}_{m, 1}\left(\boldsymbol{D}_{O}^{-1} A\right)^{m}+
\boldsymbol{\zeta}_{m, 2}\left(\boldsymbol{D}_{I}^{-1} A^{\top}\right)^{m}\right) {\mathcal{X}}_{:, p},
$
where $ p \in\{1, \cdots, D\}$,  $\boldsymbol{\zeta} \in \mathbb{R}^{M \times 2}$ denotes a trainable matrix, and $M$ signifies the number of diffusion steps. The matrices $\boldsymbol{D}_{O}^{-1} A$ and $\boldsymbol{D}_{I}^{-1} A^{\top}$ function as state transition matrices. Specifically, $\boldsymbol{D}_{\boldsymbol{O}}=\operatorname{diag}(A \mathbf{1})$ and $\boldsymbol{D}_{\boldsymbol{I}}=\operatorname{diag}(A^{\top} \mathbf{1})$ are the out-degree and in-degree diagonal matrices, respectively. Here, $\mathbf{1} \in \mathbb{R}^{N}$ is a column vector with all elements set to one.
Building upon the convolution operation, the Diffusion Convolutional Layer is defined as: $\mathcal{H}_{:, q}=\sum_{p=1}^{D} \mathcal{X}_{:, p} \star_{A} f_{\boldsymbol{\Lambda}_{q, p,:,:}} \text { for } q \in\{1, \cdots, D'\}$,
where $\mathcal{H}\in \mathbb{R}^{N \times D'}$ represents our desired output and $\boldsymbol{\Lambda} \in \mathbb{R}^{D' \times D \times M \times 2}=[\boldsymbol{\zeta}]_{q, p}$ denotes the parameter tensor.

\noindent\textbf{Predictor: }
For predictor, we utilized a single Diffusion Convolutional Layer for capturing spatial dependencies among sensors. This was followed by two 1D-Conv layers to capture traffic's temporal dynamics. Though our predictor is simplified to validate the effectiveness of our proposed framework, more advanced prediction models can be integrated. 
Within each expert, $\text{Expert}_k = ( R_k, P_k )$, 
the predictor $P_k$ takes the past $T^{\prime}$ time steps of data $x_{i, t}^\tau$ to produce its output $P_k\left(x_{i, t}^\tau\right)$.

\noindent\textbf{Reconstruction-based Gating: }
Having defined the predictors $\{P_1, P_2, ..., P_K\}$, we now describe the reconstruction-based gating mechanism that leverages the reconstructors ${\{R_1, R_2, ..., R_K\}}$, trained in the previous section, to assign weights to the the predictions of each predictor. The gating weights assigned to each predictor are defined as follows: $g_k(x_{i, w}^{\tau})=\frac{p\left({x}^{\tau}_{i,w} ; \phi_k\right)}{\sum_{k^{\prime}} p\left({x}^{\tau}_{i,w} ;  \phi_{k^{\prime}}\right)},$
Here, $x_{i, w}^{\tau}$ denotes the one-week traffic data of the sensor $i$ in task $\tau$ from the first Monday to Sunday. Moreover, $g_k(x_{i, w}^{\tau})$ represents the weight given to sensor $i$ in task $\tau$ for the predictor $P_k$. A large value of $g_k(x_{i, w}^{\tau})$ implies that the prediction from $P_k$ is particularly important for the final prediction of sensor $i$.
Utilizing these gating weights, the final prediction, which integrates the outputs from all predictors ${\{P_1, P_2, ..., P_K\}}$, is defined as follows: $\mathcal {O}\left(x_{i, t}^\tau\right)=
\sum_{k=1}^K g_k\left(x_{i,  w }^\tau\right) P_k\left(x_{i, t}^\tau\right).$
In other words, the final prediction on $x_{i, t}^\tau$ is generated by the weighted sum of the predictions made by the predictors $P_k$, and the weight is determined by how well the first week traffic data of the sensor $i$ is reconstructed by the reconstructor $R_k$, which is denoted by $g_k(x_{i, w}^{\tau})$.
Correspondingly, the loss function employed to train the entire set of experts can be formulated as follows: $\mathcal{L}_\mathcal {O}=
\sum_{i=1}^{N^\tau}\left\|\mathcal {O}\left(x_{i,t}^\tau\right)-y_{i, t}^\tau \right\|_1 .$

\noindent\textbf{Discussion: Learning on the Expanding Traffic Network. }
Real-world traffic networks expand as new sensors emerge in surrounding areas. While these sensors can reflect known traffic patterns, they often introduce new dynamics. In the first task (i.e., $\tau=1$), the entire graph $G^{1}=(V^{1}, A^{1})$ and its associated data are fully known. However, re-training the entire network for each subsequent task (i.e., $\tau>1$) is impractical. Our goal is to efficiently retain prior knowledge while accommodating new patterns via continual learning. As illustrated in Figure~\ref{fig_latent}, newly added sensors can be linked to one of the pre-existing sensor groups regardless of geographical distances. We utilize this aspect in modeling our framework.

\vspace{-1ex}
\subsubsection{Reconstruction-Based Knowledge Consolidation}
\label{sec:consoldation_loss}
\vspace{-1ex}
Inspired by the Learning without Forgetting (LwF)~\citep{cl_reg_lwf} approach, we propose a novel strategy termed the `reconstruction-based consolidation loss' to retain previously acquired knowledge while adapting to new tasks. We use $L G_k^\tau$ to denote the localized group associated with the $k$-th reconstructor for the training of task $\tau$, i.e., 
$
L G_k^\tau=\left\{\hspace{0.5mm}i \hspace{0.5mm}| \hspace{0.3mm}\underset{j}{\operatorname{argmax}} \hspace{1mm} p\left({x}^{\tau}_{i,w} ; \phi_{j}^{(\tau-1)}\right) = k , i \in \Delta V^\tau \right\},
$
where $\phi_{j}^{(\tau-1)}$ indicates the parameters of the $j$-th reconstructor that have been optimized upon completion of training up to the $(\tau-1)$-th task. A localized group collects the newly added nodes $\Delta V^\tau$ in the current task based on the reconstruction probability determined by the reconstructor that was trained and optimized in the previous task $(\tau-1)$.
The reconstruction-based consolidation loss using VAE, which incorporates the use of the localized group, is developed in a manner similar to Section~\ref{sec:training_reconstructor}. For the $k$-th expert, the marginal likelihood for all data in $LG_k^\tau$ is represented as:
$
\log p\left(x_{(L G_k^\tau[1],w)}, \ldots, x_{(L G_k^\tau\left[|LG_k^\tau|\right],w)};\phi_k\right)=\sum_{i \in L G_k^\tau} \log p\left(x_{i,w};\phi_k\right).
$
The Evidence Lower Bound (ELBO) for all experts using the variational distribution $q$ is as:
$
\mathcal{L}_{ELBO}^{LG}=
\sum_{k=1}^K 
\sum_{i \in LG_k^\tau}
\mathbb{E}_{q\left(z_k \mid x_{i,w};{\xi_k}\right)}\left[\log \frac{p\left(x_{i,w} \mid z_k;\phi_k\right)p(z_k;{\phi_k})}{q\left(z_k \mid x_{i,w};{\xi_k}\right)}\right].
$

The role of the consolidation loss is as follows: The reconstructor classifies newly added nodes $\Delta V^\tau$ of the current task into localized groups $LG^\tau_k$, based on the optimized parameters $\phi_{k}^{(\tau-1)}$ from the previous task. Then, during the learning process of reconstructor parameters $\phi_{k}^{\tau}$ in the current task $\tau$, consolidation loss strives to maintain the nodes belonging to these localized groups as much as possible. In essence, this helps preserve knowledge from the previous task while learning new information in the current task. The optimization objective for training the entire model is given by the loss function $\mathcal{L} = \mathcal{L}_\mathcal {O} - \beta\mathcal{L}_{ELBO}^{LG}$. Here, where $\beta$ is the weight of the consolidation loss. For the first task, since training is conducted on nodes partitioned into $SG_k$, the loss function is represented as $\mathcal{L} = \mathcal{L}_\mathcal {O} - \beta\mathcal{L}_{ELBO}^{SG}$.

\vspace{-1ex}
\subsubsection{Forgetting-Resilient Sampling}
\label{sec:forgetting_resilient_sampling}
\vspace{-1ex}
Next, we introduce `forgetting-resilient sampling,' which is an additional methodology that mitigates catastrophic forgetting. The core idea is to utilize the decoders of reconstructors that were trained on previous task $(\tau-1)$ to generate synthetic samples.
Within the context of our VAE-based reconstructor, for each $\text{Expert}_k$, we can sample $n_s/k$ instances of the latent variable $\{z_{k, 1}, z_{k, 2}, \ldots, z_{k, n_s/k}\} \sim p\left(z_k ; \phi_k^{(\tau-1)}\right)$.
From each $z_{k, i}$, data samples are generated according to $x_{w_{k, i}}\sim p\left(x_w \mid z_{k, i} ; \phi_k\right)$.
Thus, the dataset sampled for $\text{Expert}_k$ can be defined as $X_k^s=\left\{x_{w_{k, 1}}, x_{w_{k, 2}}, \ldots, x_{w_{k, n_s/k}}\right\}$. By aggregating over all experts, the entire sampled dataset is given by $X^s=\bigcup_{k=1}^K X_k^s$, where $|X^s| = n_s$. Note that we set $n_s$ as a hyperparameter in our model. 
The generated samples reflect the representative characteristics of each expert. Therefore, as we train our model, we integrate this generated data with the new nodes $\Delta V^\tau$ from the current task $\tau$, effectively preventing catastrophic forgetting.
Since our generated data encapsulates a week's worth of information, we have ensured synchronization of its temporal aspects when feeding it into the predictor.

An important point to stress is that, being synthetic, our generated data do not inherently have a graph structure determined by the actual geographical distances. Yet, by adopting the graph learning methodology detailed in Section \ref{sec:training_predictor}, we can adeptly address this limitation. Moreover, to efficiently preserve the existing knowledge through the sampled data, it is essential to generate data that are both similar in nature and rich in diversity. The sampling approach facilitated by our VAE guarantees efficient production of such diverse samples for each expert, which is corroborated by our experiments in  Appendix \ref{sec:generated_data}.
\vspace{-1ex}
\subsubsection{Reconstruction-Based Replay}
\label{sec:replay}
\vspace{-1ex}
In the context of continual learning, one characteristic of streaming traffic networks is that even pre-existing sensors $V^{(\tau-1)}$ display new patterns over long-term periods (e.g., several years) due to various factors such as urban development. While we strive to minimally access information from previous tasks, these sensors inherently offer indispensable information for our model training. Our main focus is on the pre-existing sensors that exhibit patterns distinct enough that they are universally unfamiliar to all experts, posing significant challenges in handling. To systematically identify these sensors, we deploy a reconstructor described as follows:
\vspace{-1pt}
\begin{equation}
\begin{aligned}
&V_R = \{{v[1]}, {v[2]}, ... , {v[n_r]}\}, \text { where } \\
&\sum_k \log p\left(x_{v[1]}; \phi_k^{(\tau-1)}\right) \leq \sum_k \log p\left(x_{v[2]}; \phi_k^{(\tau-1)}\right) \leq \dots \leq \sum_k \log p\left(x_{v[N^{(\tau-1)}]}; \phi_k^{(\tau-1)}\right) \\
&\text { and } \forall j \in \{1,2,...,N^{(\tau-1)}\}; v[j] \in V^{(\tau-1)},
\end{aligned}
\end{equation}
where $n_r$ is a hyperparameter dictating the number of elements in set $V_R$. The set $V_R$ comprises nodes, sorted in ascending order based on their reconstruction probability across all experts' reconstructors, and it is truncated to retain only the first $n_r$ entries. For training, nodes from $V_R$ are combined with the nodes of the current task, represented as $\Delta V^\tau$, along with the nodes sampled as discussed in Section ~\ref{sec:forgetting_resilient_sampling}. Consequently, the total number of nodes employed for training is given by $\Delta N^\tau + n_s + n_r$. 

\vspace{-2ex}
\section{Experiments}
\vspace{-2ex}

\noindent{\textbf{Dataset.}} Please refer to Appendix \ref{sec:dataset} for details regarding dataset.

\noindent{\textbf{Baselines.}}
Please refer to Appendix \ref{sec:baselines} for details regarding baselines.

\noindent{\textbf{Evaluation Protocol.} }
Please refer to Appendix \ref{sec:evaluation_protocol} for details regarding evaluation protocol.

\noindent{\textbf{Implementation Details.}}
Please refer to Appendix \ref{sec:implementation_details} for details regarding implementation details.

\TableMain

\begin{table*}[t]
\vspace{-2ex}
\begin{minipage}{0.41\linewidth}
  \centering
  \includegraphics[width=0.98\linewidth]{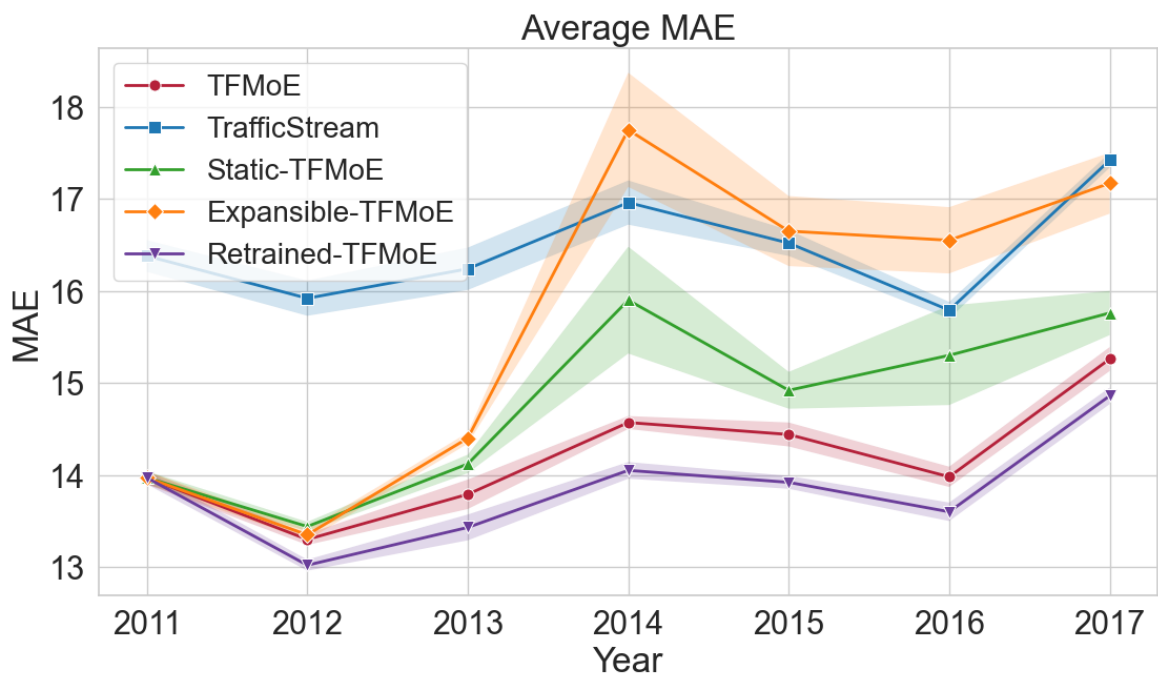} 
  \vspace{-1ex}
      \captionof{figure}{MAE of traffic flow forecasting, averaged over time horizons from 2011 to 2017, with 1-sigma error bars.}
  \label{fig_main_lineplot_mae}
\end{minipage}%
\hspace{0.015\textwidth} 
\begin{minipage}{0.59\linewidth}
  \centering
    \fontsize{7pt}{8}\selectfont
        \begin{tabular}{@{}c|c@{\hspace{5.5pt}}c@{\hspace{5.5pt}}c|c|c@{}}
        \toprule
        Model         & MAE            & RMSE           & MAPE           & \begin{tabular}[c]{@{}c@{}}Use\\ Subgraphs?\end{tabular} & \begin{tabular}[c]{@{}c@{}}Store\\ Features?\end{tabular} \\ \midrule
        Retrained-\proposed  & 13.81          & 22.97          & 18.46          & \xmark                                                        & \xmark                                                                      \\ \midrule
        Static-\proposed     & 14.81          & 24.47          & 20.78          & \xmark                                                         & \xmark                                                                      \\
        Expansible-\proposed & 15.62          & 25.69          & 22.49          & \xmark                                                         & \xmark                                                                      \\
        TrafficStream (10+$\gamma\%$)  & 16.36          & 26.73          & 23.71          & \cmark                                                        & \cmark                                                                     \\
        PECMP (10+$\gamma\%$)         & 16.02*         & 26.51*         & 22.30*         & \cmark                                                        & \cmark                                                                     \\ \midrule
        \proposed (1\%)         & \textbf{14.18} & \textbf{23.54} & \textbf{18.86} & \xmark                                                         & \xmark                                                                      \\ \bottomrule
        \end{tabular}
  \vspace{2ex}
  \captionsetup{width=1\linewidth}
  \captionof{table}{
  Prediction performance averaged over time horizons using the TrafficStream predictor across various models.  * indicates values reported from the PECMP paper.
  }
  \label{tab:60min-avg}
\end{minipage}
\vspace{-3ex}
\end{table*}

\vspace{-1ex}
\subsection{Experimental Results}
\vspace{-1ex}
Table \ref{tab:table_main} demonstrates the forecasting capabilities of each model at different time horizons (15-, 30-, and 60-minutes ahead), captured by average MAE, RMSE, and MAPE over 7 years. Figure \ref{fig_main_lineplot_mae} visualizes the MAE metrics, averaged over time horizons, for each year from 2011 to 2017. The outcomes averaged over time horizons using the TrafficStream predictor structure across various models are detailed in Table \ref{tab:60min-avg}. As the official source code of PECMP is not available, we include only comparable reported values. We have the following observations:

\protect\footnotetext{In previous studies, authors report the average metrics considering the interval of 5-minute. For instance, in the case of the 15-minute, they computed an average of predictions of 5-minute, 10-minute and 15-minute. In this work, we follow the standard practice in traffic forecasting research, for example, we compute the metric based on only the predictions made at 15-minute. This difference in calculation method may account for any discrepancies in numeric values when compared to the results of previous studies.\label{footnote1}}

\noindent\textbf{1)} Static-\proposed~ underperforms because it uses only 2011 data for predictions up to 2017. This highlights the necessity of integrating new data for accurate forecasting.
\noindent\textbf{2)} Results from Expansible-\proposed~ reveal that depending solely on newly added sensor data each year, without referencing historical knowledge, degrades performance due to catastrophic forgetting.
\noindent\textbf{3)} Existing models (i.e., TrafficStream and PECMP) connect newly added nodes with pre-existing nodes by constructing subgraphs around them, while \proposed~does not require an access to any pre-existing nodes thanks to the graph structure learning. 
Besides, existing models also construct subgraphs around replayed nodes, which further adds the number of accessed pre-exisitng nodes (denoted by $\gamma$ \footnote{Our empirical observations indicate that for models employing subgraphs, the average access rate to pre-existing sensors exceeds 20\%, implying $\gamma\% > 20\%$. This is because replayed nodes also possess subgraphs.\label{footnote2}}), while~\proposed~merely accesses the replayed nodes without construction of subgraphs. Note that existing models use 10\% of the number of pre-existing sensors for replay, while~\proposed~only uses 1\%.
In short,~\proposed~outperforms existing models even with a significantly limited access to pre-existing nodes.
\noindent\textbf{4)} To detect and replay sensors with patterns that are either similar or different between the current and the previous year, existing models typically require storing features from all sensors of the previous year in a separate memory. However, as elaborated in Section~\ref{sec:replay}, our approach uses a VAE-based reconstructor, allowing us to analyze and compare with past patterns using only the data from the current task. This eliminates the need for dedicated memory storage for historical data. From the perspective of continual learning, which aims to minimize memory usage for past tasks, this presents a significant advantage. 
\noindent\textbf{5)} While other methods employ various strategies against catastrophic forgetting, \proposed~ stands superior, accessing only 1\% of the pre-existing nodes corresponding to past tasks. This signifies its robustness and suitability for real-world traffic forecasting.

\begin{table*}[t]
\vspace{-2ex}
\caption{Component analysis of ~\proposed.}
\fontsize{8pt}{9}\selectfont
\begin{adjustbox}{center}
\begin{tabular}{@{}c|ccc|ccc|ccc|cll@{}}
\toprule
\multirow{2}{*}{Method} & \multicolumn{3}{c|}{15 min} & \multicolumn{3}{c|}{30 min} & \multicolumn{3}{c|}{60 min} & \multicolumn{3}{c}{\multirow{2}{*}{\begin{tabular}[c]{@{}c@{}}Training Time\\ (sec)\end{tabular}}} \\ \cmidrule(lr){2-10}
                        & MAE     & RMSE    & MAPE    & MAE     & RMSE    & MAPE    & MAE     & RMSE    & MAPE    & \multicolumn{3}{c}{}                                                                               \\ \midrule
w/o Consol              & 13.41   & 21.67   & 18.16   & 15.13   & 24.52   & 20.46   & 19.03   & 30.57   & 25.28   & \multicolumn{3}{c}{229}                                                                            \\
w/o Sampling            & 12.84   & 20.97   & 17.16   & 14.44   & 23.68   & 18.86   & 18.14   & 29.53   & 23.31   & \multicolumn{3}{c}{230}                                                                            \\
w/o Replay              & 13.11   & 21.36   & 18.86   & 14.72   & 24.19   & 20.38   & 18.47   & 30.41   & 25.93   & \multicolumn{3}{c}{233}                                                                            \\
\proposed                   & \textbf{12.48} & \textbf{20.48} & \textbf{16.72} & \textbf{13.93} & \textbf{23.00} & \textbf{18.13} & \textbf{17.20} & \textbf{28.36} & \textbf{22.17}   & \multicolumn{3}{c}{244}                                                                            \\ \bottomrule
\end{tabular}
\end{adjustbox}
% \vspace{-2ex}
\label{tab:componet_analysis}
\vspace{-2ex}
\end{table*}

\vspace{-1ex}
\subsection{Model Analysis}
\vspace{-1ex}
\noindent\textbf{Component Analysis:}
We delve deeply into the individual components of \proposed. To systematically evaluate the contribution of each component to the overall performance, we introduce four model variations: (i)~\proposed~ which integrates all the components, (ii) w/o Consol which excludes the Consolidation Loss (detailed in Section~\ref{sec:consoldation_loss}), (iii) w/o Sampling which operates without the Sampling mechanism (detailed in Section~\ref{sec:forgetting_resilient_sampling}), and (iv) w/o Replay that does not employ the Replay strategy (detailed in Section~\ref{sec:replay}). The comparative performances of these models are presented in Table \ref{tab:componet_analysis}.
We observe that each component plays a pivotal role in enhancing the model's efficacy, as elucidated in the respective sections. Notably,~\proposed, with its full suite of components, stands out as the top-performing model, validating the efficacy of our proposed methodologies.

\begin{wrapfigure}{R}{0.6\textwidth} 
\vspace{-1ex}
  \centering
  \includegraphics[width=0.6\textwidth]{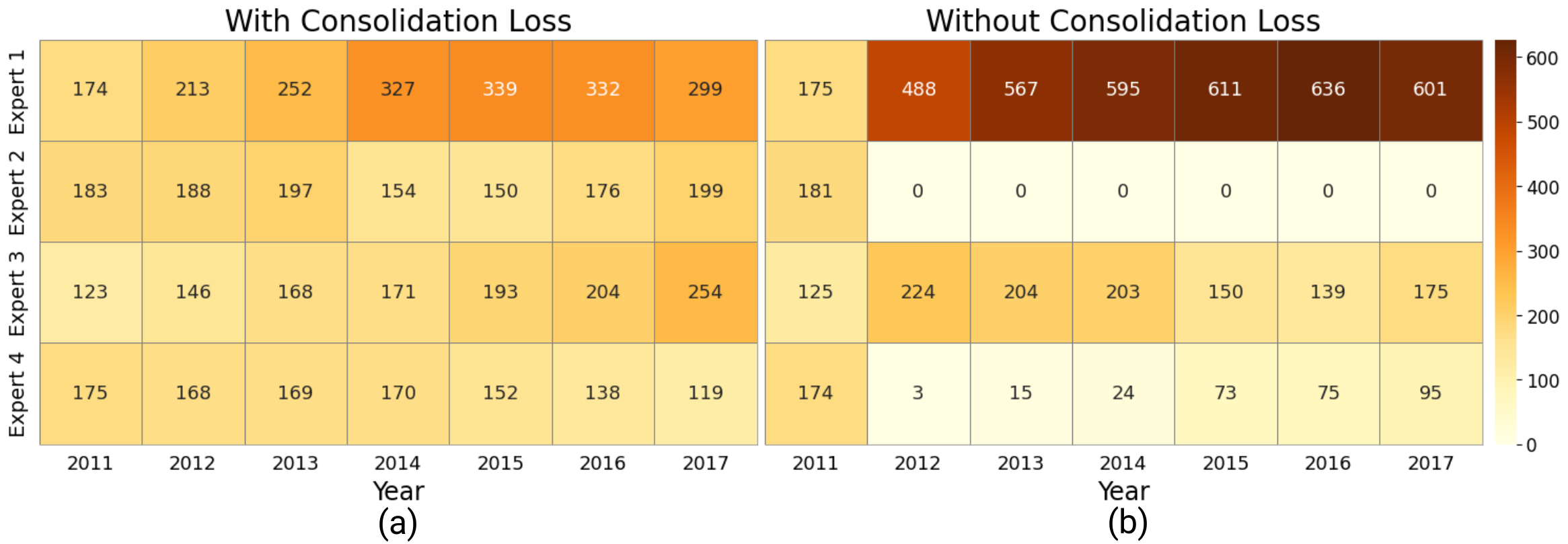} 
  \vspace{-2ex}
  \caption{Comparison of sensor allocation to experts over the years (2011-2017) (a) With consolidation loss and (b) Without consolidation loss.}
  \vspace{-1ex}
  \label{fig_heatmap}
\end{wrapfigure} 

\smallskip
\noindent\textbf{Effect of Consolidation Loss on Expert Utilization: }
In this section, we evaluate the impact of the consolidation loss described in Section~\ref{sec:consoldation_loss} on expert utilization, emphasizing its role in addressing catastrophic forgetting.
Our ablation study, depicted in Figure \ref{fig_heatmap}, employs two heatmaps. 
The $x$-axis spans the years, and the $y$-axis represents our four experts. 
Each cell in the heatmap indicates the number of sensors allocated to an expert during training, based on the lowest reconstruction error.
Figure \ref{fig_heatmap}~(a) and (b) contrast scenarios with and without the consolidation loss, respectively. We observe that incorporating the consolidation loss results in a balanced sensor distribution across experts throughout the years, which helps address catastrophic forgetting by maintaining each expert's unique contributions. Without the consolidation loss, sensors predominantly cluster around one expert after the first year, indicating a failure in diverse expert utilization and the onset of catastrophic forgetting.
In essence, the consolidation loss is pivotal for balanced expert engagement in \proposed, preventing dominance by any single expert and fostering effective continual learning.

\vspace{-0.3ex}
\section{CONCLUSION}
\vspace{-0.3ex}
In this paper, we introduced the Traffic Forecasting Mixture of Experts (\proposed), a novel continual learning approach designed specifically for long-term streaming networks. Informed by real-world traffic patterns,~\proposed~generates a specialized expert model for each homogeneous group, effectively adapting to evolving traffic networks and their inherent complexities. 
To overcome the significant obstacle of catastrophic forgetting in continual learning scenarios, we introduce three complementary mechanisms: `Reconstruction-Based Knowledge Consolidation Loss', `Forgetting-Resilient Sampling', and `Reconstruction-Based Replay mechanisms', which allow~\proposed~to retain essential prior knowledge effectively while seamlessly assimilating new information.
The merit of our approach is validated through extensive experiments on a real-world long-term streaming network dataset, PEMSD3-Stream. Not only did~\proposed~demonstrate superior performance in traffic flow forecasting, but it also showcased resilience against catastrophic forgetting, a key factor in the continual learning of traffic flow forecasting in long-term streaming networks. As such, our model offers a potent and efficient approach to tackling the evolving challenges of traffic flow forecasting.

% \section*{References}
\clearpage

\bibliographystyle{plain}
\bibliography{sample_base}

\begin{thebibliography}{10}

\bibitem{agcrn}
Lei Bai, Lina Yao, Can Li, Xianzhi Wang, and Can Wang.
\newblock Adaptive graph convolutional recurrent network for traffic forecasting.
\newblock {\em NeurIPS}, 2020.

\bibitem{gcn3}
Joan Bruna, Wojciech Zaremba, Arthur Szlam, and Yann LeCun.
\newblock Spectral networks and locally connected networks on graphs.
\newblock {\em arXiv preprint arXiv:1312.6203}, 2013.

\bibitem{stemgnn}
Defu Cao, Yujing Wang, Juanyong Duan, Ce~Zhang, Xia Zhu, Congrui Huang, Yunhai Tong, Bixiong Xu, Jing Bai, Jie Tong, et~al.
\newblock Spectral temporal graph neural network for multivariate time-series forecasting.
\newblock {\em NeurIPS}, 2020.

\bibitem{random_replay_1}
Arslan Chaudhry, Marcus Rohrbach, Mohamed Elhoseiny, Thalaiyasingam Ajanthan, Puneet~K Dokania, Philip~HS Torr, and Marc'Aurelio Ranzato.
\newblock On tiny episodic memories in continual learning.
\newblock {\em arXiv preprint arXiv:1902.10486}, 2019.

\bibitem{pems}
Chao Chen, Karl Petty, Alexander Skabardonis, Pravin Varaiya, and Zhanfeng Jia.
\newblock Freeway performance measurement system: mining loop detector data.
\newblock {\em Transportation Research Record}, 2001.

\bibitem{trafficstream}
Xu~Chen, Junshan Wang, and Kunqing Xie.
\newblock Trafficstream: A streaming traffic flow forecasting framework based on graph neural networks and continual learning.
\newblock {\em arXiv preprint arXiv:2106.06273}, 2021.

\bibitem{zgcnnet}
Yuzhou Chen, Ignacio Segovia, and Yulia~R Gel.
\newblock Z-gcnets: Time zigzags at graph convolutional networks for time series forecasting.
\newblock In {\em ICML}. PMLR, 2021.

\bibitem{gru}
Junyoung Chung, Caglar Gulcehre, KyungHyun Cho, and Yoshua Bengio.
\newblock Empirical evaluation of gated recurrent neural networks on sequence modeling.
\newblock {\em arXiv preprint arXiv:1412.3555}, 2014.

\bibitem{gcn2}
Micha{\"e}l Defferrard, Xavier Bresson, and Pierre Vandergheynst.
\newblock Convolutional neural networks on graphs with fast localized spectral filtering.
\newblock {\em NeurIPS}, 29, 2016.

\bibitem{stgode}
Zheng Fang, Qingqing Long, Guojie Song, and Kunqing Xie.
\newblock Spatial-temporal graph ode networks for traffic flow forecasting.
\newblock In {\em KDD}, 2021.

\bibitem{cluster2}
Kan Guo, Yongli Hu, Yanfeng Sun, Sean Qian, Junbin Gao, and Baocai Yin.
\newblock Hierarchical graph convolution network for traffic forecasting.
\newblock In {\em Proceedings of the AAAI conference on artificial intelligence}, volume~35, pages 151--159, 2021.

\bibitem{astgcn}
Shengnan Guo, Youfang Lin, Ning Feng, Chao Song, and Huaiyu Wan.
\newblock Attention based spatial-temporal graph convolutional networks for traffic flow forecasting.
\newblock In {\em AAAI}, 2019.

\bibitem{moe}
Robert~A Jacobs, Michael~I Jordan, Steven~J Nowlan, and Geoffrey~E Hinton.
\newblock Adaptive mixtures of local experts.
\newblock {\em Neural computation}, 3(1):79--87, 1991.

\bibitem{gumbel2}
Eric Jang, Shixiang Gu, and Ben Poole.
\newblock Categorical reparameterization with gumbel-softmax.
\newblock {\em arXiv preprint arXiv:1611.01144}, 2016.

\bibitem{adam}
Diederik~P Kingma and Jimmy Ba.
\newblock Adam: A method for stochastic optimization.
\newblock {\em arXiv preprint arXiv:1412.6980}, 2014.

\bibitem{vae}
Diederik~P Kingma and Max Welling.
\newblock Auto-encoding variational bayes.
\newblock {\em arXiv preprint arXiv:1312.6114}, 2013.

\bibitem{gcn}
Thomas~N Kipf and Max Welling.
\newblock Semi-supervised classification with graph convolutional networks.
\newblock {\em arXiv preprint arXiv:1609.02907}, 2016.

\bibitem{cl_reg_ewc}
James Kirkpatrick, Razvan Pascanu, Neil Rabinowitz, Joel Veness, Guillaume Desjardins, Andrei~A Rusu, Kieran Milan, John Quan, Tiago Ramalho, Agnieszka Grabska-Barwinska, et~al.
\newblock Overcoming catastrophic forgetting in neural networks.
\newblock {\em Proceedings of the national academy of sciences}, 114(13):3521--3526, 2017.

\bibitem{dstagnn}
Shiyong Lan, Yitong Ma, Weikang Huang, Wenwu Wang, Hongyu Yang, and Pyang Li.
\newblock Dstagnn: Dynamic spatial-temporal aware graph neural network for traffic flow forecasting.
\newblock In {\em International Conference on Machine Learning}, pages 11906--11917. PMLR, 2022.

\bibitem{moe_generative}
Soochan Lee, Junsoo Ha, Dongsu Zhang, and Gunhee Kim.
\newblock A neural dirichlet process mixture model for task-free continual learning.
\newblock {\em arXiv preprint arXiv:2001.00689}, 2020.

\bibitem{stfgnn}
Mengzhang Li and Zhanxing Zhu.
\newblock Spatial-temporal fusion graph neural networks for traffic flow forecasting.
\newblock In {\em AAAI}, 2021.

\bibitem{expert_traffic_1}
Shuhao Li, Yue Cui, Yan Zhao, Weidong Yang, Ruiyuan Zhang, and Xiaofang Zhou.
\newblock St-moe: Spatio-temporal mixture-of-experts for debiasing in traffic prediction.
\newblock In {\em Proceedings of the 32nd ACM International Conference on Information and Knowledge Management}, pages 1208--1217, 2023.

\bibitem{dcrnn}
Yaguang Li, Rose Yu, Cyrus Shahabi, and Yan Liu.
\newblock Diffusion convolutional recurrent neural network: Data-driven traffic forecasting.
\newblock {\em arXiv preprint arXiv:1707.01926}, 2017.

\bibitem{cl_reg_lwf}
Zhizhong Li and Derek Hoiem.
\newblock Learning without forgetting.
\newblock {\em IEEE transactions on pattern analysis and machine intelligence}, 40(12):2935--2947, 2017.

\bibitem{stag-gcn}
Bin Lu, Xiaoying Gan, Haiming Jin, Luoyi Fu, and Haisong Zhang.
\newblock Spatiotemporal adaptive gated graph convolution network for urban traffic flow forecasting.
\newblock In {\em CIKM}, 2020.

\bibitem{gumbel}
Chris~J Maddison, Andriy Mnih, and Yee~Whye Teh.
\newblock The concrete distribution: A continuous relaxation of discrete random variables.
\newblock {\em arXiv preprint arXiv:1611.00712}, 2016.

\bibitem{cl_arc_packnet}
Arun Mallya and Svetlana Lazebnik.
\newblock Packnet: Adding multiple tasks to a single network by iterative pruning.
\newblock In {\em Proceedings of the IEEE conference on Computer Vision and Pattern Recognition}, pages 7765--7773, 2018.

\bibitem{st-gart}
Cheonbok Park, Chunggi Lee, Hyojin Bahng, Yunwon Tae, Seungmin Jin, Kihwan Kim, Sungahn Ko, and Jaegul Choo.
\newblock St-grat: A novel spatio-temporal graph attention networks for accurately forecasting dynamically changing road speed.
\newblock In {\em CIKM}, 2020.

\bibitem{cl_mem_icarl}
Sylvestre-Alvise Rebuffi, Alexander Kolesnikov, Georg Sperl, and Christoph~H Lampert.
\newblock icarl: Incremental classifier and representation learning.
\newblock In {\em Proceedings of the IEEE conference on Computer Vision and Pattern Recognition}, pages 2001--2010, 2017.

\bibitem{cl_mem_er}
Anthony Robins.
\newblock Catastrophic forgetting, rehearsal and pseudorehearsal.
\newblock {\em Connection Science}, 7(2):123--146, 1995.

\bibitem{cl_arc_pnn}
Andrei~A Rusu, Neil~C Rabinowitz, Guillaume Desjardins, Hubert Soyer, James Kirkpatrick, Koray Kavukcuoglu, Razvan Pascanu, and Raia Hadsell.
\newblock Progressive neural networks.
\newblock {\em arXiv preprint arXiv:1606.04671}, 2016.

\bibitem{gts}
Chao Shang, Jie Chen, and Jinbo Bi.
\newblock Discrete graph structure learning for forecasting multiple time series.
\newblock {\em arXiv preprint arXiv:2101.06861}, 2021.

\bibitem{cl_generative_1}
Hanul Shin, Jung~Kwon Lee, Jaehong Kim, and Jiwon Kim.
\newblock Continual learning with deep generative replay.
\newblock {\em Advances in neural information processing systems}, 30, 2017.

\bibitem{stsgcn}
Chao Song, Youfang Lin, Shengnan Guo, and Huaiyu Wan.
\newblock Spatial-temporal synchronous graph convolutional networks: A new framework for spatial-temporal network data forecasting.
\newblock In {\em AAAI}, 2020.

\bibitem{cluster_student}
Laurens Van~der Maaten and Geoffrey Hinton.
\newblock Visualizing data using t-sne.
\newblock {\em Journal of machine learning research}, 9(11), 2008.

\bibitem{gat}
Petar Veli{\v{c}}kovi{\'c}, Guillem Cucurull, Arantxa Casanova, Adriana Romero, Pietro Lio, and Yoshua Bengio.
\newblock Graph attention networks.
\newblock {\em arXiv preprint arXiv:1710.10903}, 2017.

\bibitem{random_replay_2}
Jeffrey~S Vitter.
\newblock Random sampling with a reservoir.
\newblock {\em ACM Transactions on Mathematical Software (TOMS)}, 11(1):37--57, 1985.

\bibitem{cl_traffic_1}
Binwu Wang, Yudong Zhang, Jiahao Shi, Pengkun Wang, Xu~Wang, Lei Bai, and Yang Wang.
\newblock Knowledge expansion and consolidation for continual traffic prediction with expanding graphs.
\newblock {\em IEEE Transactions on Intelligent Transportation Systems}, 2023.

\bibitem{pecpm}
Binwu Wang, Yudong Zhang, Xu~Wang, Pengkun Wang, Zhengyang Zhou, Lei Bai, and Yang Wang.
\newblock Pattern expansion and consolidation on evolving graphs for continual traffic prediction.
\newblock In {\em Proceedings of the 29th ACM SIGKDD Conference on Knowledge Discovery and Data Mining}, pages 2223--2232, 2023.

\bibitem{cl_graph_4}
Chen Wang, Yuheng Qiu, Dasong Gao, and Sebastian Scherer.
\newblock Lifelong graph learning.
\newblock In {\em Proceedings of the IEEE/CVF conference on computer vision and pattern recognition}, pages 13719--13728, 2022.

\bibitem{expert_traffic_2}
Hongjun Wang, Jiyuan Chen, Zipei Fan, Zhiwen Zhang, Zekun Cai, and Xuan Song.
\newblock St-expertnet: A deep expert framework for traffic prediction.
\newblock {\em IEEE Transactions on Knowledge and Data Engineering}, 2022.

\bibitem{cl_graph_3}
Junshan Wang, Guojie Song, Yi~Wu, and Liang Wang.
\newblock Streaming graph neural networks via continual learning.
\newblock In {\em Proceedings of the 29th ACM international conference on information \& knowledge management}, pages 1515--1524, 2020.

\bibitem{cl_generative_2}
Chenshen Wu, Luis Herranz, Xialei Liu, Joost Van De~Weijer, Bogdan Raducanu, et~al.
\newblock Memory replay gans: Learning to generate new categories without forgetting.
\newblock {\em Advances in Neural Information Processing Systems}, 31, 2018.

\bibitem{wavenet}
Zonghan Wu, Shirui Pan, Guodong Long, Jing Jiang, and Chengqi Zhang.
\newblock Graph wavenet for deep spatial-temporal graph modeling.
\newblock {\em arXiv preprint arXiv:1906.00121}, 2019.

\bibitem{cluster_dec}
Junyuan Xie, Ross Girshick, and Ali Farhadi.
\newblock Unsupervised deep embedding for clustering analysis.
\newblock In {\em International conference on machine learning}, pages 478--487. PMLR, 2016.

\bibitem{cluster1}
Qinge Xie, Tiancheng Guo, Yang Chen, Yu~Xiao, Xin Wang, and Ben~Y Zhao.
\newblock Deep graph convolutional networks for incident-driven traffic speed prediction.
\newblock In {\em Proceedings of the 29th ACM international conference on information \& knowledge management}, pages 1665--1674, 2020.

\bibitem{rnncnn2}
Huaxiu Yao, Xianfeng Tang, Hua Wei, Guanjie Zheng, Yanwei Yu, and Zhenhui Li.
\newblock Modeling spatial-temporal dynamics for traffic prediction.
\newblock {\em arXiv preprint arXiv:1803.01254}, 2018.

\bibitem{rnncnn3}
Huaxiu Yao, Fei Wu, Jintao Ke, Xianfeng Tang, Yitian Jia, Siyu Lu, Pinghua Gong, Jieping Ye, and Zhenhui Li.
\newblock Deep multi-view spatial-temporal network for taxi demand prediction.
\newblock In {\em AAAI}, 2018.

\bibitem{stgcn}
Bing Yu, Haoteng Yin, and Zhanxing Zhu.
\newblock Spatio-temporal graph convolutional networks: A deep learning framework for traffic forecasting.
\newblock {\em arXiv preprint arXiv:1709.04875}, 2017.

\bibitem{cl_reg_si}
Friedemann Zenke, Ben Poole, and Surya Ganguli.
\newblock Continual learning through synaptic intelligence.
\newblock In {\em International conference on machine learning}, pages 3987--3995. PMLR, 2017.

\bibitem{rnncnn}
Junbo Zhang, Yu~Zheng, Dekang Qi, Ruiyuan Li, Xiuwen Yi, and Tianrui Li.
\newblock Predicting citywide crowd flows using deep spatio-temporal residual networks.
\newblock {\em Artificial Intelligence}, 2018.

\bibitem{gman}
Chuanpan Zheng, Xiaoliang Fan, Cheng Wang, and Jianzhong Qi.
\newblock Gman: A graph multi-attention network for traffic prediction.
\newblock In {\em AAAI}, 2020.

\bibitem{cl_graph_1}
Fan Zhou and Chengtai Cao.
\newblock Overcoming catastrophic forgetting in graph neural networks with experience replay.
\newblock In {\em Proceedings of the AAAI Conference on Artificial Intelligence}, volume~35, pages 4714--4722, 2021.

\bibitem{gsl}
Yanqiao Zhu, Weizhi Xu, Jinghao Zhang, Yuanqi Du, Jieyu Zhang, Qiang Liu, Carl Yang, and Shu Wu.
\newblock A survey on graph structure learning: Progress and opportunities.
\newblock {\em arXiv e-prints}, pages arXiv--2103, 2021.

\end{thebibliography}

%%%%%%%%%%%%%%%%%%%%%%%%%%%%%%%%%%%%%%%%%%%%%%%%%%%%%%%%%%%%

\clearpage
\appendix
\linespread{1} 
\section*{Appendix}

\section{Dataset}
\label{sec:dataset}
In order to evaluate the performance of \proposed, we conduct experiments using the PEMSD3-Stream dataset, which is a real-world highway traffic dataset collected by the Caltrans Performance Measurement System (PeMS) in real time every 30 seconds~\citep{pems}. The traffic data are aggregated into 5-minute intervals, resulting in 12 time steps per hour. The PEMSD3-Stream dataset consists of traffic flow information in the North Central Area from 2011 to 2017. Consistent with previous studies, we select data from July 10th to August 9th annually. The input data are rescaled using Z-score normalization. The PEMSD3-Stream traffic network is continually expanding, meaning that sensors installed in the $\tau$-th year remain operational in subsequent years.
The recorded feature in this dataset is solely traffic flow, resulting in a homogenous graph structure. It is important to note that, in contrast to previous studies that employed pre-defined graph structures based on geographical distances, our approach adopts a graph learning strategy incorporating synthetic nodes sampled by a Variational Autoencoder (VAE). A detailed summary of the datasets, including edge information based on geographical distances that define the pre-defined graph structure, can be found in Table~\ref{tab:pre_defined_graph}.

\begin{table}[h]
\caption{Dataset statistics of pre-defined graph structure.}
\vspace{1.5ex}
\begin{adjustbox}{center}
\begin{tabular}{@{}c|cccccccl@{}}
\toprule
\quad Year \quad     & 2011 & 2012 & 2013 & 2014 & 2015 & 2016 & 2017 &  \\ \midrule
\quad \# Nodes \quad & 655  & 715  & 786  & 822  & 834  & 850  & 871  &  \\ \midrule
\quad \# Edges \quad & 1577 & 1929 & 2316 & 2536 & 2594 & 2691 & 2788 &  \\ \bottomrule
\end{tabular}
\end{adjustbox}
\label{tab:pre_defined_graph}
\end{table}

\section{Baselines}
\label{sec:baselines}
We compare~\proposed~with the following baselines:

\begin{itemize}[leftmargin=5mm]

\item \textbf{GRU~\citep{gru}:} GRU operates by leveraging gated recurrent units, which enable it to effectively manage sequential data by adapting the information flow through its internal memory mechanism. For the training of this GRU, we utilize data from every node each year.

\item \textbf{DCRNN~\citep{dcrnn}:} DCRNN integrates both spatial and temporal dependency by using diffusion convolution and encoder-decoder architecture. For the training of this DCRNN, we utilize data from every node each year.

\item \textbf{STSGCN~\citep{stsgcn}:} STSGCN employs a localized spatial-temporal graph convolutional module, which allows the model to directly and concurrently capture intricate localized spatial-temporal correlations. For the training of this STSGCN, we utilize data from every node each year.

\item \textbf{TrafficStream~\citep{trafficstream}:} The first work that applies continual learning techniques to a streaming traffic dataset. In order to mitigate catastrophic forgetting, TrafficStream proposes the use of Elastic Weight Consolidation (EWC) and a replay mechanism utilizing Jensen-Shannon divergence.

\item \textbf{PECPM~\citep{pecpm}:} PECPM utilizes a bank module and Elastic Weight Consolidation (EWC) for pattern storage in evolving traffic networks, allowing the model to adapt seamlessly with pattern expansion and consolidation.

\item \textbf{Retrained-\proposed:} We retrain~\proposed~every year with all nodes given in each year.
We initialize each year's model with the parameters learned from the previous year's model. This approach serves as an upper bound in the context of continual learning.

% \looseness=-1
\item \textbf{Static-\proposed:} We train~\proposed~solely on data from the first year (2011) and then directly use the trained model, without further training, to predict the traffic flow of all subsequent years.

\item \textbf{Expansible-\proposed:} We train~\proposed~in an online manner each year, utilizing only data from the newly added sensors, while initializing each year's model with the parameters learned from the previous year's model. This model is equivalent to~\proposed~without sampling and replay, as described in Sections~\ref{sec:forgetting_resilient_sampling} and~\ref{sec:replay}, respectively.

\end{itemize}
\clearpage

\section{Evaluation Protocol}
\label{sec:evaluation_protocol}

We leverage three standard performance metrics for model evaluation: mean absolute error (MAE), root mean squared error (RMSE), and mean absolute percentage error (MAPE). The datasets are divided into training, validation, and test sets with a distribution ratio of 6:2:2. Utilizing datasets aggregated into 5-minute intervals, our model uses one hour of historical data (12 time steps) to make predictions for the subsequent hour (12 time steps). 
Experiments are conducted on 13th Gen Intel(R) Core(TM) i9-13900K and NVIDIA GeForce RTX 4090.
We conducted each experiment five times using different seeds and report the mean performance. In the line plot, we have illustrated 1-sigma error bars. 
\vspace{3ex}

\section{Implementation Details}
\label{sec:implementation_details}
The detailed hyperparameter setting of~\proposed~is as follows: The number of clusters $K$ (i.e., Experts) is tuned in $\{1, 2, 3, 4, 5, 6\}$. In the `Reconstruction-Based Clustering' (Section~\ref{sec:reconstruction_based_clustering}), an AutoEncoder structure is adopted for the pre-training reconstructor, where both the encoder and decoder are designed with three MLPs each. The coefficient for the clustering loss, $\alpha$, is tuned in $\{1e-5, 1e-4, 1e-3, 1e-2\}$. In the `Training Reconstructor' (Section~\ref{sec:training_reconstructor}), the reconstructor is based on a Variational AutoEncoder (VAE) architecture, with both the encoder and decoder again made up of three MLPs. For both the AutoEncoder and VAE, the encoded latent hidden dimension is tuned in $\{4, 8, 16, 32, 64\}$.  In the `Training Predictor of Expert' (Section~\ref{sec:training_predictor}), the predictor comprises of one GNN layer and two 1-D convolution layers, with hidden dimension tuned in $\{4, 8, 16, 32, 64\}$. 
The diffusion convolution operation's diffusion step is set to $M=1$. The consolidation loss, $\beta$, for the `Reconstruction-Based Knowledge Consolidation' (Section~\ref{sec:consoldation_loss}) is tuned in $\{0.01, 0.1, 1, 10\}$. In the `Forgetting-Resilient Sampling' (Section~\ref{sec:forgetting_resilient_sampling}), the number of samples, $n_s$, is set to 9\% of the current task graph size, while in the `Reconstruction-Based Replay' (Section~\ref{sec:replay}), the number of replays, $n_r$, is set to 1\% of the current task graph size.
Training was executed with 50 epochs for the first task and 10 epochs for the following tasks. We employed a batch size of 128 and utilized the Adam~\citep{adam} optimizer, setting learning rates of 0.001, 0.0001, and 0.01 for the pre-training reconstructor, reconstructor, and predictor, respectively.
\vspace{3ex}

\section{Visualization of Reconstructor and Predictor Outputs}
\vspace{2ex}

\begin{figure}[h]
\begin{center}
\includegraphics[width=1.05\linewidth]{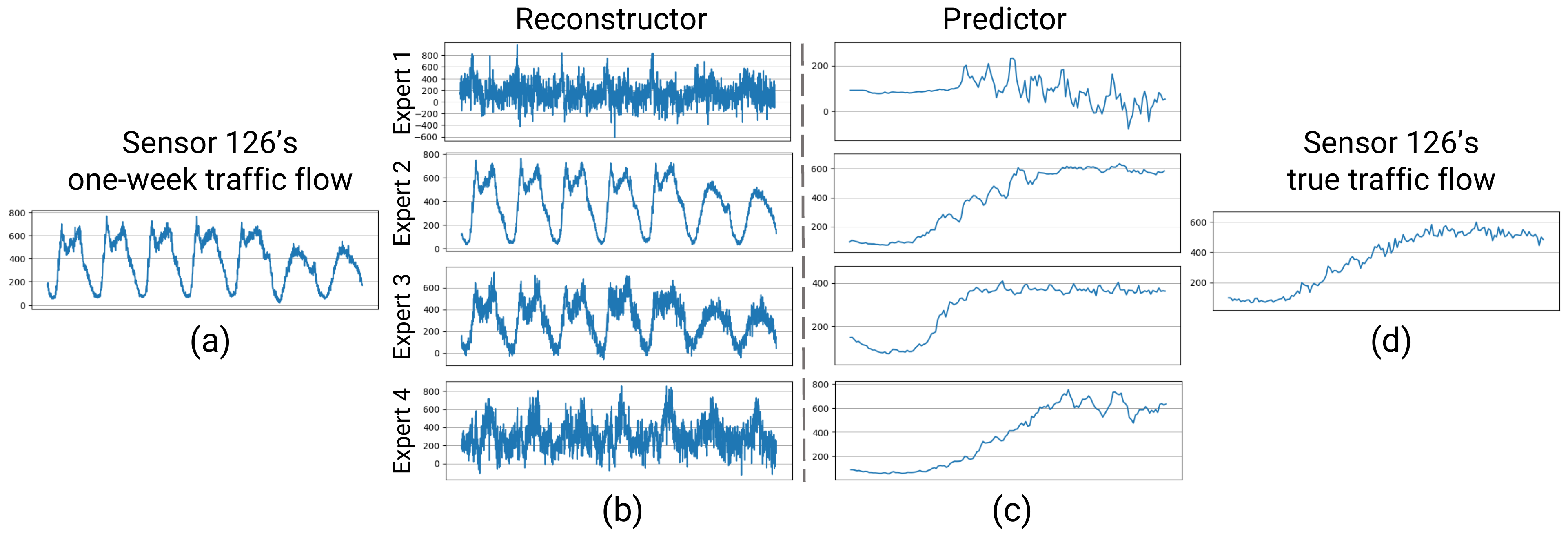}
\end{center}
\caption{(a) Actual one-week traffic flow data measured from sensor 126 in the year 2012. (b) One-week traffic flow reconstructed by each expert's Reconstructor. (c) Traffic flow forecasted by each expert's Predictor. (d) Actual traffic flow data measured from sensor 126.}
\label{fig_expert_visual}
\end{figure}

In this section, we perform detailed analysis and visualization of ~\proposed~ performance, focusing specifically on individual outputs from each expert. We have the following observations in Figure~\ref{fig_expert_visual}:
\textbf{1)} Upon close inspection of the one-week traffic flow data from sensor 126 (i.e. Figure~\ref{fig_expert_visual}~(a)), it becomes evident that expert 2 has managed to achieve the most accurate reconstruction. 
\textbf{2)} Moreover, a detailed examination of the traffic flows predicted by each expert's Predictor (i.e., Figure~\ref{fig_expert_visual}~(c)) demonstrates that expert 2’s predictor closely mirrors the true traffic flow (i.e., Figure~\ref{fig_expert_visual}~(d)). 
\textbf{3)} While the predicted traffic flow from predictor 2 and 3 may seem similar, a closer observation of the scale on the y-axis reveals that the output from expert 2’s predictor aligns more precisely with the true traffic data.
Through the above comprehensive analyses, we visually underscore the notion that the experts proficient in reconstruction also tend to provide more accurate predictions. This in-depth evaluation helps elucidate the functioning of the components of~\proposed~and their individual contributions to the overall performance.

\vspace{3ex}

\section{Visualization of Synthetic Data Generated through Decoders}
\label{sec:generated_data}
\vspace{3ex}

\begin{figure}[h]
\begin{center}
\includegraphics[width=1.0\linewidth]{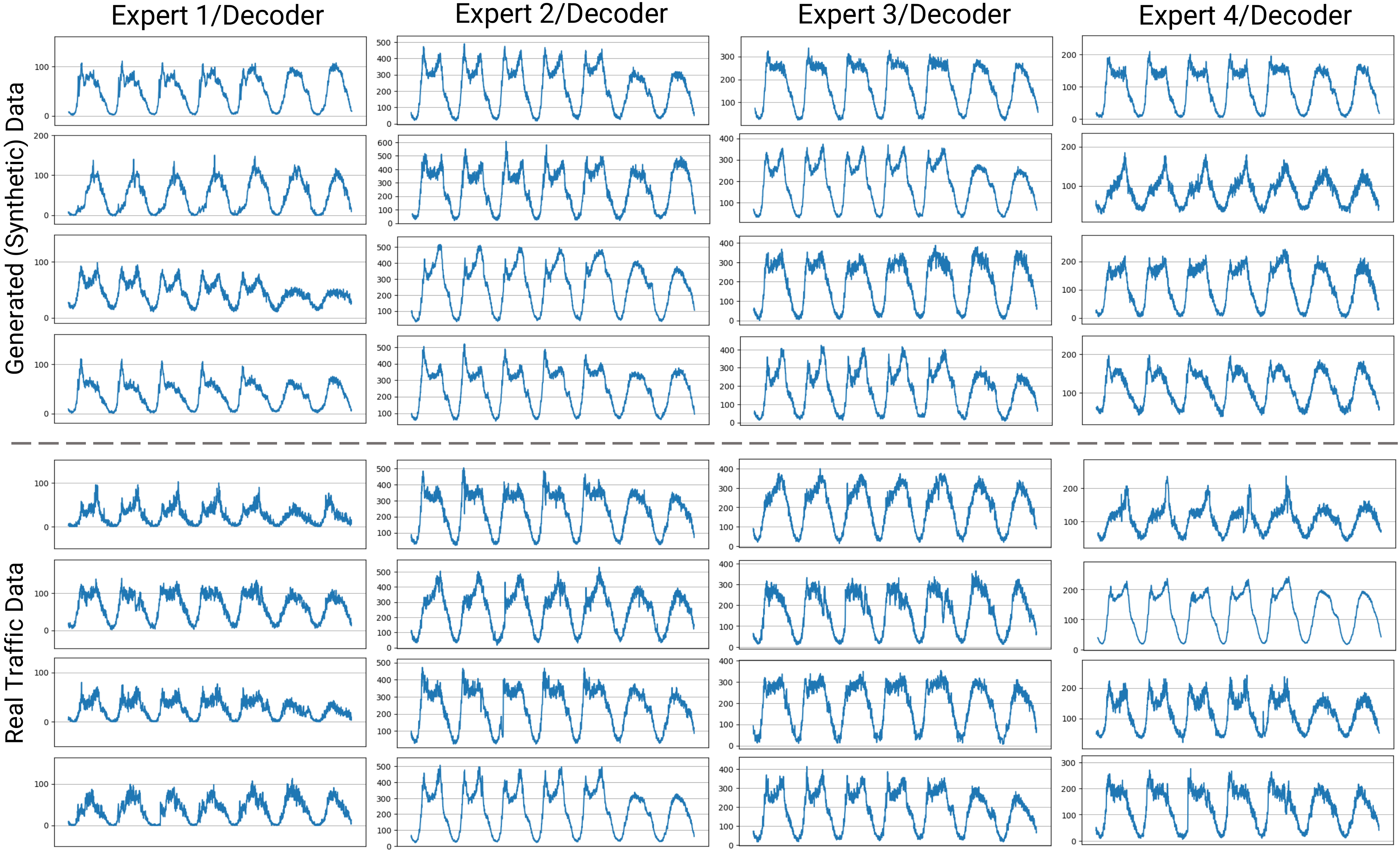}
\end{center}
\vspace{3ex}
\caption{Visualization of synthetic data generated through decoders. Above the dashed line, we present plots of synthetic samples drawn using $x_{w}\sim p\left(x_w \mid z_{k} ; \phi_k\right)$, where $z_{k} \sim p\left(z_k ; \phi_k\right)$ for each Expert. Below the dashed line, the plots showcase the actual traffic data assigned to the respective Expert.
}
\label{fig_generated_data}
\vspace{3ex}
\end{figure}

In order to investigate how well the synthetic data sampling process, described in Section \ref{sec:forgetting_resilient_sampling}, captures the characteristics of each Expert's real traffic data, we visualized samples generated by the VAE decoder of each Expert in Figure \ref{fig_generated_data}. Specifically, synthetic samples were drawn using $x_{w}\sim p\left(x_w \mid z_{k} ; \phi_k\right)$, where $z_{k} \sim p\left(z_k ; \phi_k\right)$. A comparison between the real traffic data and the synthetic data in Figure \ref{fig_generated_data} reveals that the synthesized data indeed reflects the individual characteristics of each Expert's real traffic data. This is evident both from the scale of the y-axis and the overall shape of the plots. As mentioned in Section \ref{sec:forgetting_resilient_sampling}, our goal is to generate data that is not only similar in nature but also rich in diversity.  The synthetic data showcased in  Figure \ref{fig_generated_data}~aligns well with this objective.

\clearpage

\section{Optimizing Expert Selection}
\vspace{-1ex}

\begin{table*}[h]
\caption{Performance metrics (MAE, RMSE, MAPE) and training time over different numbers of experts.}
\fontsize{8pt}{7}\selectfont
\begin{adjustbox}{center}
\begin{tabular}{@{}c|ccc|ccc|ccc|cll@{}}
\toprule
\multirow{2}{*}{\# of Expert} & \multicolumn{3}{c|}{15 min} & \multicolumn{3}{c|}{30 min} & \multicolumn{3}{c|}{60 min} & \multicolumn{3}{c}{\multirow{2}{*}{\begin{tabular}[c]{@{}c@{}}Training Time\\ (sec)\end{tabular}}} \\ \cmidrule(lr){2-10}
                              & MAE     & RMSE    & MAPE    & MAE     & RMSE    & MAPE    & MAE     & RMSE    & MAPE    & \multicolumn{3}{c}{}                                                                               \\ \midrule
K = 1                         & 13.43   & 22.23   & 17.75   & 15.37   & 25.71   & 19.89   & 20.36   & 33.63   & 25.60   & \multicolumn{3}{c}{61}                                                                             \\
K = 2                         & 13.01   & 21.34   & 17.59   & 14.75   & 24.36   & 19.62   & 19.08   & 31.22   & 24.66   & \multicolumn{3}{c}{115}                                                                            \\
K = 3                         & 12.76   & 20.91   & 17.13   & 14.31   & 23.55   & 18.86   & 18.04   & 29.43   & 23.68   & \multicolumn{3}{c}{177}                                                                            \\
K = 4                         & 12.48   & 20.48   & 16.72   & 13.93   & 23.00   & 18.13   & 17.20   & 28.36   & 22.17   & \multicolumn{3}{c}{244}                                                                            \\
K = 5                         & 12.49   & 20.51   & 16.53   & 13.95   & 23.06   & 18.22   & 17.16   & 28.27   & 22.31   & \multicolumn{3}{c}{297}                                                                            \\
K = 6                         & 12.42   & 20.35   & 16.47   & 13.86   & 22.84   & 18.13   & 17.11   & 28.00   & 21.79   & \multicolumn{3}{c}{363}                                                                            \\ \bottomrule
\end{tabular}
\end{adjustbox}
\label{tab:num_expert}
\end{table*}

A critical hyperparameter in our model is the number of experts. To thoroughly examine the impact of number of experts on model performance, we conducted additional experiments, the results of which are presented in Table~\ref{tab:num_expert}. Our findings indicate that as the number of experts increases, there is a corresponding improvement in prediction accuracy. This enhancement can be attributed to the specialization of each expert in distinct data clusters, thereby enhancing the model's overall accuracy. However, this benefit comes with a trade-off: an increase in the number of experts leads to longer training times. It is important to note that the performance improvement begins to plateau when the number of experts is increased from four to five; this increment does not yield a significant enhancement in error metrics but incurs additional computational expenses. Thus, four experts seem to offer an optimal balance between accuracy and computational efficiency.

The observed diminishing returns beyond a certain number of experts suggest potential redundancy, characterized by overlapping functionalities among the experts. To address this challenge, we propose several strategies aimed at optimizing expert selection. An empirical approach involves incrementally adding experts and monitoring the resultant changes in performance. Alternatively, visual inspection tools such as t-SNE, as depicted in Figure~\ref{fig_latent}, can offer valuable insights. More sophisticated methods may leverage our generator model to identify experts producing similar samples, indicating redundancy.

Building upon the previous discussion, we can consider a dynamic expert adjustment strategy that could be applicable both in selecting the initial number of experts and throughout the continual learning process. By utilizing the $LG$, outlined in Section~\ref{sec:consoldation_loss}, we can identify experts with low utilization rates. These underutilized experts, who may not contribute significantly to performance enhancement, could be candidates for removal or for merging with other experts exhibiting similar characteristics. On the other hand, when an expert's scope extends over a diverse dataset, there is an opportunity for refinement. In such cases, we recommend performing clustering on the latent vectors associated with this expert to segregate the data into more distinct subsets. Subsequently, we can assign dedicated experts to each of these clustered groups, ensuring more focused and specialized learning.

\section{Replay Method Analysis: Reconstruction-Based Replay vs Random Sampling}
\vspace{-1ex}

\begin{table*}[h]
\caption{Comparison of performance metrics (MAE, RMSE, MAPE) between our Reconstruction-Based Replay and the random sampling strategy for replay methods.}
\fontsize{8pt}{7}\selectfont
\begin{adjustbox}{center}
\begin{tabular}{@{}c|c|ccc|ccc|ccc@{}}
\toprule
\multirow{2}{*}{Replay Method}   & \multirow{2}{*}{Replay Ratio} & \multicolumn{3}{c|}{15 min} & \multicolumn{3}{c|}{30 min} & \multicolumn{3}{c}{60 min} \\ \cmidrule(l){3-11} 
                                 &                               & MAE     & RMSE    & MAPE    & MAE     & RMSE    & MAPE    & MAE     & RMSE    & MAPE   \\ \midrule
\multirow{3}{*}{Random}          & 1\%                           & 13.07   & 21.31   & 18.68   & 14.65   & 24.07   & 20.22   & 18.35   & 30.24   & 25.62  \\
                                 & 3\%                           & 13.02   & 21.22   & 18.49   & 14.61   & 23.97   & 20.10   & 18.25   & 30.03   & 25.59  \\
                                 & 5\%                          & 12.86   & 21.05   & 17.98   & 14.41   & 23.72   & 19.62   & 17.93   & 29.61   & 24.62  \\ \midrule
\multirow{3}{*}{Ours}            & 1\%                           & 12.48   & 20.48   & 16.72   & 13.93   & 23.00   & 18.13   & 17.20   & 28.36   & 22.17  \\
                                 & 3\%                           & 12.48   & 20.46   & 16.75   & 13.95   & 22.98   & 18.18   & 17.28   & 28.34   & 22.33  \\
                                 & 5\%                          & 12.50   & 20.49   & 16.61   & 13.98   & 23.01   & 18.15   & 17.36   & 28.44   & 22.41  \\ \bottomrule
\end{tabular}
\end{adjustbox}
\label{tab:reconsruction_based_replay}
\end{table*}

In the realm of continual learning, the efficiency of the random sampling method has been substantiated through numerous studies~\citep{random_replay_1,random_replay_2}. This section presents a comparative analysis between our Reconstruction-Based Replay methodology, as detailed in Section \ref{sec:replay}, and the random sampling approach. The results of this experimental comparison are presented in Table~\ref{tab:reconsruction_based_replay}. The `Replay Ratio' in the table indicates the proportion of data replayed relative to the size of the current task's graph. Notably, our method demonstrates superior performance compared to random sampling across all replay ratios. Interestingly, in our approach, increasing the replay ratio does not significantly enhance performance, suggesting that our method adeptly identifies and replays pivotal samples for the current task with minimal data. This characteristic is particularly advantageous from the perspective of continual learning, where the primary goal is to minimize access to pre-existing nodes from prior tasks. In contrast, the performance of random sampling shows a direct correlation with increased replay ratios. This trend can be ascribed to its underlying mechanism, which involves randomly selecting samples, thereby necessitating a higher replay ratio to achieve comparable performance enhancements.

\section{Evaluating the Impact of Sampling Ratio on Model Performance}

\begin{table*}[h]
\caption{Comparison of performance metrics (MAE, RMSE, MAPE) across varying sampling ratios.}
\fontsize{8pt}{7}\selectfont
\begin{adjustbox}{center}
\begin{tabular}{@{}c|ccc|ccc|ccc|c@{}}
\toprule
\multirow{2}{*}{Sampling Ratio} & \multicolumn{3}{c|}{15 min} & \multicolumn{3}{c|}{30 min} & \multicolumn{3}{c|}{60 min} & \multirow{2}{*}{\begin{tabular}[c]{@{}c@{}}Training Time\\ (sec)\end{tabular}} \\ \cmidrule(lr){2-10}
                                & MAE     & RMSE    & MAPE    & MAE     & RMSE    & MAPE    & MAE     & RMSE    & MAPE    &                                                                                \\ \midrule
0\%                             & 12.84   & 20.97   & 17.16   & 14.44   & 23.68   & 18.86   & 18.14   & 29.53   & 23.31   & 230                                                                            \\
5\%                             & 12.53   & 20.54   & 16.70   & 14.01   & 23.08   & 18.22   & 17.33   & 28.41   & 22.30   & 236                                                                            \\
10\%                            & 12.48   & 20.47  & 16.64   & 13.93   & 22.98   & 18.11   & 17.19   & 28.30   & 22.16   & 244                                                                            \\
20\%                            & 12.46   & 20.45   & 16.85   & 13.92   & 22.96   & 18.23   & 17.21   & 28.38   & 22.22   & 262                                                                            \\
30\%                            & 12.46   & 20.48   & 16.83   & 13.93   & 23.01   & 18.27   & 17.18   & 28.41   & 22.29   & 291                                                                            \\ \bottomrule
\end{tabular}
\end{adjustbox}
\label{tab:sampling_ablation}
\end{table*}

In this section, building upon the Forgetting-Resilient Sampling method introduced in Section~\ref{sec:forgetting_resilient_sampling}, we delve deeper into the influence of the sampling ratio on model performance. Our findings are tabulated in Table~\ref{tab:sampling_ablation}, where the `Sampling Ratio' denotes the proportion of data sampled relative to the size of the current task's graph. Throughout our experiments, we maintained a consistent replay ratio of 1\%. Our empirical results highlight the critical role of the sampling method. Specifically, completely abstaining from sampling led to the poorest performance, while an increase in the sampling ratio to 10\% resulted in the best performance improvements. Implementing sampling is crucial, as it provides each Expert with representative data that solidifies their foundational knowledge, effectively counteracting catastrophic forgetting. The reason for the sampling ratio needs to be higher than the replay ratio stems from the distinct requirements of replay and sampling; while replay focuses on pinpointing universally unfamiliar sensors across all experts, sampling demands a proportionate amount of samples corresponding to the number of Experts, as each Expert requires distinct representative samples. Furthermore, our results suggest that a moderate sampling ratio suffices to safeguard the memory of each Expert, precluding the need for an excessively large pool of samples. 

\section{Analysis of the Effects of Clustering in the Pre-training Stage}

\begin{table*}[h]
\caption{Comparison of performance between the Random-Cluster model and the~\proposed~ model.}
\fontsize{9pt}{8}\selectfont
\begin{adjustbox}{center}
\begin{tabular}{@{}c|ccc|ccc|ccc@{}}
\toprule
\multirow{2}{*}{Method} & \multicolumn{3}{c|}{15 min} & \multicolumn{3}{c|}{30 min} & \multicolumn{3}{c}{60 min} \\ \cmidrule(l){2-10} 
                        & MAE     & RMSE    & MAPE    & MAE     & RMSE    & MAPE    & MAE     & RMSE    & MAPE   \\ \midrule
Random-Cluster          & 12.84   & 21.06   & 17.35   & 14.51   & 23.90   & 19.26   & 18.66   & 30.51   & 24.22  \\
\proposed                       & 12.48   & 20.48   & 16.72   & 13.93   & 23.00   & 18.13   & 17.20   & 28.36   & 22.17  \\ \bottomrule
\end{tabular}
\end{adjustbox}
\label{table:random_cluster}
\end{table*}

In this section, we delve into the impact of clustering during the pre-training stage on model performance. We contrast our method with the `Random-Cluster' model, wherein nodes are arbitrarily assigned to clusters. As illustrated in Table \ref{table:random_cluster}, our model significantly surpasses the Random-Cluster model in performance. In the Random-Cluster model, nodes are randomly allocated to clusters, leading each Expert to handle a subset of data, which inherently reflects the overall node distribution. The formation of $K$ clusters using Random-Cluster method and the integration of their results resembles a form of ensemble learning, with each expert trained on same distributions that collectively approximate the entire node distribution.

As discussed in Section~\ref{sec:intro}, our proposed approach involves segmenting traffic data into multiple homogeneous groups, with a dedicated Expert assigned to each. This contrasts with the Random-Cluster model, which does not differentiate within the data spectrum. Our method ensures that each Expert is specialized in the data distribution of their assigned group. The introduction of a new node is handled by the most relevant Expert, thus minimizing the impact on other Experts and aiding in preventing catastrophic forgetting. Additionally, each Expert's VAE reconstructor more effectively generates and reconstructs the data distribution of its specific group, and the predictors are more accurate in forecasting for their respective group distributions. This strategic design underpins the superior performance of our model over the Random-Cluster model.

\section{Optimizing Temporal Data Ranges in Autoencoder Models for Traffic Pattern Analysis.}

\begin{table*}[h]
\caption{Performance comparison of Autoencoder models using different data periods: one week, two weeks, only Monday, and Friday \& Saturday, as inputs. }
\fontsize{9pt}{8}\selectfont
\begin{adjustbox}{center}
\begin{tabular}{@{}c|ccc|ccc|ccc@{}}
\toprule
\multirow{2}{*}{Method} & \multicolumn{3}{c|}{15 min} & \multicolumn{3}{c|}{30 min} & \multicolumn{3}{c}{60 min} \\ \cmidrule(l){2-10} 
                        & MAE     & RMSE    & MAPE    & MAE     & RMSE    & MAPE    & MAE     & RMSE    & MAPE   \\ \midrule
Two week              & 12.46   & 20.52   & 16.61   & 13.92   & 23.04   & 18.23   & 17.25   & 28.42   & 22.00  \\
One week              & 12.48   & 20.48   & 16.72   & 13.93   & 23.00   & 18.13   & 17.20   & 28.36   & 22.17  \\ 
Friday \& Saturday    & 12.60   & 20.66   & 16.73   & 14.11   & 23.25   & 18.40   & 17.57   & 28.76   & 22.53  \\ 
Monday                & 12.71   & 20.78   & 16.98   & 14.24   & 23.35   & 18.86   & 17.67   & 28.84   & 23.31  \\ \bottomrule
\end{tabular}
\end{adjustbox}
\label{table:one_week_reson}
\end{table*}

In our modeling process, we use \( x^{\tau}_{i,w} \in \mathbb{R}^{week} \), which denotes the initial one-week traffic data for the \(i^{th}\) node, selected from the first full week of data (Monday to Sunday) in the training dataset. The decision to utilize one week of data was made to capture a comprehensive range of patterns, encompassing both weekdays and weekends, since traffic patterns in weekdays significantly differ from weekends. Moreover, we considered the variance within weekdays (between Monday and Friday) and weekends (Saturday \& Sunday), underlining the need for a holistic weekly dataset. To validate our selection of one-week data, we conducted additional experiments. 

In Table~\ref{table:one_week_reson}, `Friday \& Saturday' uses data from Friday and Saturday instead of a full week in the autoencoder, while `Monday' uses only Monday's data. Similarly, `Two week' refers to the use of two weeks' data. Results show `Monday' has the lowest performance; `Friday \& Saturday' which captures both weekday and weekend patterns, shows better performance. `One week' and `Two week' models, which capture both weekday and weekend patterns and consider the variance within weekdays and weekends, show the best performance. The similar performance of `One week' and `Two week' demonstrates that using only one week of data is sufficient for effectively clustering the sensors according to their characteristics.

\section{Complexity Analysis}
\label{sec:complexity_analysis}
\looseness=-1
\noindent{\textbf{Time Complexity:}} The time complexity of our model is primarily dominated by the computational requirements of the predictor's Graph Neural Network operations in the Localized Adaptation Stage, which are $O(N^2MK)$. Here, $N$ represents the number of nodes, $K$ is the number of clusters, and $M$ denotes the diffusion step. 

\smallskip
\noindent{\textbf{Space Complexity:}} The space complexity of our model is predominantly governed by the $O(N^2K)$ complexity arising from the graph learning process of the predictor in the Localized Adaptation Stage. 

\looseness=-1
It is important to note that the calculated time complexity of $O(N^2MK)$ and space complexity of $O(N^2K)$ impose a burden only on the initial task, as they utilize the entire given dataset. As we mentioned in our introduction, the reason we adopt continual learning is to prevent the forgetting of existing knowledge while training only on newly added data, rather than retraining on the entire dataset. In practice, the volume of new data added each year is typically much smaller than the existing nodes. After the initial task, training is conducted using a number of nodes proportional to the newly added nodes, $\Delta N=\left|\Delta V\right|$. Additionally, as described in Section~\ref{sec:implementation_details}, we set the diffusion step $M=1$. Consequently, both the time complexity and space complexity after the first year become $O({\Delta N}^2K)$.

\section{Limitation}
\label{sec:limitation}
One limitation of our study arises from the scarcity of datasets and baseline models in the field of continual traffic forecasting, an area that is relatively nascent and not yet extensively explored. Despite this challenge, our extensive visualizations and experiments have validated that our model is exceptionally well-suited and effectively designed for continual traffic forecasting, demonstrating significant performance improvements over existing models. We anticipate that as continual traffic forecasting gains prominence and research in this domain intensifies, there will be an influx of more diverse datasets and baseline models. Such advancements will not only invigorate the field but also pave the way for more comprehensive future studies and deeper comparative analyses.

\section{Future Work}
Although our approach emphasizes continual learning to avoid retraining on the entire dataset, focusing instead on new data, the initial task still requires training on the full dataset.
As detailed in Section~\ref{sec:complexity_analysis}, this leads to a complexity of $O(N^2K)$, presenting challenges when the initial task involves a large number of nodes. 
In situations where similar data forms densely packed, extensive sensor groups, one feasible solution might involve sampling a suitable number of nodes within each cluster for training. In another scenario, if clustering results in numerous sensor groups, each containing a moderate amount of data, a viable approach could be to divide the initial task into several smaller subtasks. Each subtask would consist of fewer nodes, and we could apply continual learning techniques within these subtasks.

The methodologies considered for applying to large datasets could potentially become new research topics within the `Large-Scale Dataset Learning' or `Distributed Computing' research areas. However, these extend beyond the scope of our current research, which is focused on continual learning methodologies. Therefore, we will leave these considerations as future work.

\clearpage
\section{Algorithm Description}
\label{sec:algorithm_description}

\begin{algorithm*}[h]
\caption{Traffic Forecasting Mixture of Experts (\textsc{TFMoE})}
\begin{algorithmic}[1]
\Require Training data for every task $ \mathcal{D}^\tau = \{D^\tau_i \}_i= \{y^{\tau}_{i,t}, x^{\tau}_{i,t},x^{\tau}_{i,w}\}_{(i,t)}$ , nodes for every task $V^\tau$, weight of clustering loss $\alpha$ and consolidation loss $\beta$, hyperparameters $n_s$ and $n_r$
\Function{Train TFMoE}{}
    \State Pre-training reconstructor $R_k$ of each Expert using \Call{Pre-TrainingStage}{$\mathbf{X}^1_w = \{x^{1}_{i,w}\}_i$}
    \For{every task $\tau$}
        \If{$\tau = 1$}
            \State Train Experts $(R_k, P_k)$ on first task data $\mathcal{D}^1$ with loss $\mathcal{L}=\mathcal{L}_{\mathcal{O}}-
\beta\mathcal{L}^{SG}_{ELBO}$
        \Else
            \State Synthetic data samples $D^s \gets$ \Call{ForgettingResilientSampling}{$\phi^{(\tau-1)},n_s$}
            \State Important nodes $V_R \gets$ \Call{ReconstructionBasedReplay}{$V^{(\tau-1)}, \phi^{(\tau-1)},n_r$}
            \State Aggregate data for current task: $D^* \gets D^s \cup \left\{D_i^\tau \mid i \in \Delta V^\tau \cup V_R\right\}$
            \State Construct $LG_k^\tau$ on $D^*$
            \State Train Experts $(R_k, P_k)$ on aggregated data $D^*$ with loss $\mathcal{L}=\mathcal{L}_{\mathcal{O}}-
\beta\mathcal{L}^{LG}_{ELBO}$
        \EndIf
    \EndFor
\EndFunction
\end{algorithmic}
\end{algorithm*}

\begin{algorithm*}[h]
\caption{\textsc{Pre-Training Stage}}
\begin{algorithmic}[1]
\Require First task data $\mathbf{X}^1_w$
    \State Train autoencoder $R_{pre}$ with loss $\mathcal{L}_{recon}$
    \State Get $K$ cluster centroids $\left[\mu_1 ; \ldots ; \mu_K\right]$ using latent vectors $\kappa_i=R_{pre;enc}(x^1_{i,w})$
    \State Train autoencoder $R_{pre}$ with loss $\mathcal{L}_p=\mathcal{L}_{recon}+\alpha\mathcal{L}_{cluster}$ 
    \State Construct $SG_k$ utilizing cluster assignment probability
    \State Train each reconstructor $R_k$ with loss $\mathcal{L}^{SG}_{ELBO}$ based on cluster $SG_k$
    \State \Return pre-trained reconstructor $R_k$

\end{algorithmic}
\end{algorithm*}

\begin{algorithm*}[h]
\caption{\textsc{Forgetting-Resilient Sampling}}
\begin{algorithmic}[1]
\Require $\phi^{(\tau-1)}$, $n_s$
    \For{each Expert $k$}
        \State Sample latent variables: $z_{k,i} \sim p(z_k; \phi_k^{(\tau-1)})$ for $i = 1$ to $n_s/k$
        \State Generate data samples: $x_{w_{k,i}} \sim p (x_w | z_{k,i}; \phi_k)$
        \State Aggregate samples: $X^s_k \gets \{x_{w_{k,1}}, x_{w_{k,2}}, \dots, x_{w_{k,n_s/k}}\}$
    \EndFor
    \State Aggregated dataset: $X^s \gets \bigcup^K_{k=1} X_k^s$
    \State Synchronize aggregated dataset: $D^s\gets \text{Sync}(X^s)$
    \State \Return $D^s$
    
\end{algorithmic}
\end{algorithm*}

\begin{algorithm*}[!h]
\caption{\textsc{Reconstruction-Based Replay}}
\begin{algorithmic}[1]
\Require $V^{(\tau-1)}, \phi^{(\tau-1)}$, $n_r$
    % \State Rank nodes by reconstruction probability $\Sigma_{k} \log p(x_{i,w};\phi_k^{(\tau-1)})$
    % \State Select $V_R$ with top $n_r$ ranked nodes from node set $V^{(\tau-1)}$
    \State Rank nodes by reconstruction probability $\Sigma_{k} \log p(x_{i,w};\phi_k^{(\tau-1)})$ in ascending order
    \State Select the first $n_r$ nodes to form $V_R$ from node set $V^{(\tau-1)}$
    % \State Select the top $n_r$ ranked nodes from $V^{(\tau-1)}$ and assign to $V_R$
    
    \State \Return $V_R$
\end{algorithmic}
\end{algorithm*}

\clearpage

\section{Hyperparameter Sensitivity Analysis}
\label{sec:hyperparameter_sensitivity}

\begin{table*}[!ht]
\caption{Sensitivity analysis of reconstructor hidden dimension on model performance.}
\fontsize{9}{8}\selectfont
\begin{adjustbox}{center}
\begin{tabular}{@{\hspace{10pt}}c@{\hspace{10pt}}|ccc|ccc|ccc@{}}
\toprule
\multirow{2}{*}{\begin{tabular}[c]{@{}c@{}}Reconstructor\\Hidden Dimensions\end{tabular}} & \multicolumn{3}{c|}{15 min} & \multicolumn{3}{c|}{30 min} & \multicolumn{3}{c}{60 min} \\ \cmidrule(l){2-10} 
                             & MAE     & RMSE    & MAPE    & MAE     & RMSE    & MAPE    & MAE     & RMSE    & MAPE   \\ \midrule
dim = 4                        & 12.67   & 20.75   & 16.95   & 14.25   & 23.46   & 18.71   & 17.78   & 29.11   & 22.68  \\
dim = 8                        & 12.59   & 20.64   & 16.74   & 14.07   & 23.19   & 18.35   & 17.46   & 28.68   & 22.48  \\
dim = 16                        & 12.64   & 20.68   & 16.80   & 14.18   & 23.29   & 18.61   & 17.71   & 28.92   & 22.81  \\
dim = 32                        & 12.48   & 20.48   & 16.72   & 13.93   & 23.00   & 18.13   & 17.20   & 28.36   & 22.17  \\
dim = 64                        & 12.68   & 20.80   & 16.69   & 14.13   & 23.32   & 18.44   & 17.52   & 28.84   & 22.56  \\ \bottomrule
\end{tabular}
\end{adjustbox}
\end{table*}
\vspace{5ex}

\begin{table*}[!ht]
\caption{Sensitivity analysis of predictor hidden dimension on model performance.}
\fontsize{9}{8}\selectfont
\begin{adjustbox}{center}
\begin{tabular}{@{\hspace{10pt}}c@{\hspace{10pt}}|ccc|ccc|ccc@{}}
\toprule
\multirow{2}{*}{\begin{tabular}[c]{@{}c@{}}Predictor\\Hidden Dimensions\end{tabular}} & \multicolumn{3}{c|}{15 min} & \multicolumn{3}{c|}{30 min} & \multicolumn{3}{c}{60 min} \\ \cmidrule(l){2-10} 
                             & MAE     & RMSE    & MAPE    & MAE     & RMSE    & MAPE    & MAE     & RMSE    & MAPE   \\ \midrule
dim = 4                        & 13.22   & 21.70   & 17.51   & 14.83   & 24.46   & 19.39   & 18.76   & 30.67   & 24.05  \\
dim = 8                        & 12.89   & 21.16   & 17.18   & 14.41   & 23.74   & 19.10   & 18.13   & 29.64   & 23.76  \\
dim = 16                        & 12.74   & 20.93   & 16.98   & 14.25   & 23.48   & 18.86   & 17.79   & 29.17   & 22.92  \\
dim = 32                        & 12.48   & 20.48   & 16.72   & 13.93   & 23.00   & 18.13   & 17.20   & 28.36   & 22.17  \\
dim = 64                        & 12.52   & 20.35   & 16.33   & 13.96   & 22.85   & 17.94   & 17.16   & 28.00   & 22.03  \\ \bottomrule
\end{tabular}
\end{adjustbox}
\end{table*}
\vspace{5ex}

\begin{table*}[!ht]
\caption{Sensitivity analysis of clustering loss weight (i.e., \(\alpha\)) on model performance.}
\fontsize{9}{8}\selectfont
\begin{adjustbox}{center}
\begin{tabular}{@{\hspace{10pt}}c@{\hspace{10pt}}|ccc|ccc|ccc@{}}
\toprule
\multirow{2}{*}{\begin{tabular}[c]{@{}c@{}}Clustering\\Loss Weight\end{tabular}} & \multicolumn{3}{c|}{15 min} & \multicolumn{3}{c|}{30 min} & \multicolumn{3}{c}{60 min} \\ \cmidrule(l){2-10} 
                             & MAE     & RMSE    & MAPE    & MAE     & RMSE    & MAPE    & MAE     & RMSE    & MAPE   \\ \midrule
$\alpha$ = 0                 & 12.54   & 20.61   & 16.67   & 14.01   & 23.15   & 18.25   & 17.29   & 28.45   & 22.38  \\
$\alpha$ = 1e-5           & 12.54   & 20.54   & 16.74   & 14.01   & 23.08   & 18.26   & 17.30   & 28.34   & 22.40  \\
$\alpha$ = 1e-4            & 12.48   & 20.48   & 16.72   & 13.93   & 23.00   & 18.13   & 17.20   & 28.36   & 22.17  \\
$\alpha$ = 1e-3             & 12.58   & 20.64   & 16.70   & 14.07   & 23.20   & 18.32   & 17.48   & 28.65   & 22.42  \\
$\alpha$ = 1e-2            & 12.60   & 20.63   & 17.04   & 14.04   & 23.18   & 18.38   & 17.38   & 28.56   & 22.51  \\ \bottomrule
\end{tabular}
\end{adjustbox}
\end{table*}

\vspace{5ex}

\begin{table*}[!ht]
\caption{Sensitivity analysis of consolidation loss weight (i.e., \(\beta\)) on model performance.}
\fontsize{9}{8}\selectfont
\begin{adjustbox}{center}
\begin{tabular}{@{\hspace{10pt}}c@{\hspace{10pt}}|ccc|ccc|ccc@{}}
\toprule
\multirow{2}{*}{\begin{tabular}[c]{@{}c@{}}Consolidation\\Loss Weight\end{tabular}} & \multicolumn{3}{c|}{15 min} & \multicolumn{3}{c|}{30 min} & \multicolumn{3}{c}{60 min} \\ \cmidrule(l){2-10} 
                             & MAE     & RMSE    & MAPE    & MAE     & RMSE    & MAPE    & MAE     & RMSE    & MAPE   \\ \midrule
$\beta$ = 0                 & 13.41   & 21.67   & 18.16   & 15.13   & 24.52   & 20.46   & 19.03   & 30.57   & 25.28  \\
$\beta$ = 0.01              & 12.63   & 20.66   & 17.02   & 14.08   & 23.15   & 18.44   & 17.41   & 28.45   & 22.47  \\
$\beta$ = 0.1               & 12.48   & 20.48   & 16.72   & 13.93   & 23.00   & 18.13   & 17.20   & 28.36   & 22.17  \\
$\beta$ = 1                 & 12.53   & 20.59   & 16.54   & 14.05   & 23.19   & 18.30   & 17.47   & 28.71   & 22.59  \\
$\beta$ = 10                & 12.62   & 20.71   & 16.64   & 14.11   & 23.26   & 18.17   & 17.46   & 28.65   & 22.34  \\ \bottomrule
\end{tabular}
\end{adjustbox}
\end{table*}
\clearpage

\begin{table*}[!ht]
\caption{Sensitivity analysis of learning rate on model performance. The table presents the performance metrics for different combinations of learning rates, where `r\_lr' stands for reconstructor learning rate and `p\_lr' stands for predictor learning rate .}
\fontsize{9}{8}\selectfont
\begin{adjustbox}{center}
\begin{tabular}{@{\hspace{10pt}}c@{\hspace{10pt}}|ccc|ccc|ccc@{}}
\toprule
\multirow{2}{*}{[r\_lr, p\_lr]} & \multicolumn{3}{c|}{15 min} & \multicolumn{3}{c|}{30 min} & \multicolumn{3}{c}{60 min} \\ \cmidrule(l){2-10} 
                             & MAE     & RMSE    & MAPE    & MAE     & RMSE    & MAPE    & MAE     & RMSE    & MAPE   \\ \midrule
$[5e-4, 1e-2]$               & 12.68   & 20.80   & 16.66   & 14.14   & 23.29   & 18.30   & 17.57   & 28.79   & 22.44  \\
$[5e-4, 5e-3]$              & 12.70   & 20.83   & 16.91   & 14.15   & 23.27   & 18.41   & 17.39   & 28.50   & 22.30  \\
$[1e-4, 5e-2]$               & 13.26   & 21.60   & 17.98   & 14.95   & 24.51   & 19.87   & 19.11   & 31.13   & 25.01  \\
$[1e-4, 1e-2]$               & 12.48   & 20.48   & 16.72   & 13.93   & 23.00   & 18.13   & 17.20   & 28.36   & 22.17  \\
$[1e-4, 5e-3]$               & 12.63   & 20.67   & 16.79   & 14.21   & 23.31   & 18.65   & 17.69   & 28.81   & 22.88  \\
$[5e-5, 5e-2]$               & 13.55   & 22.13   & 18.19   & 15.31   & 25.03   & 20.35   & 19.55   & 31.76   & 25.06  \\
$[5e-5, 1e-2]$               & 12.63   & 20.67   & 17.04   & 14.19   & 23.34   & 18.70   & 17.82   & 29.18   & 23.18  \\
$[5e-5, 5e-3]$               & 12.61   & 20.63   & 16.98   & 14.14   & 23.28   & 18.37   & 17.70   & 28.96   & 22.83  \\ \bottomrule
\end{tabular}
\end{adjustbox}
\end{table*}

\vspace{5ex}

\begin{table*}[!ht]
\caption{Sensitivity analysis of batch size on model performance.}
\fontsize{9}{8}\selectfont
\begin{adjustbox}{center}
\begin{tabular}{@{\hspace{10pt}}c@{\hspace{10pt}}|ccc|ccc|ccc@{}}
\toprule
\multirow{2}{*}{Batch Size} & \multicolumn{3}{c|}{15 min} & \multicolumn{3}{c|}{30 min} & \multicolumn{3}{c}{60 min} \\ \cmidrule(l){2-10} 
                             & MAE     & RMSE    & MAPE    & MAE     & RMSE    & MAPE    & MAE     & RMSE    & MAPE   \\ \midrule
size = 16                 & 12.55   & 20.57   & 16.59   & 14.00   & 23.02   & 18.23   & 17.27   & 28.22   & 22.28  \\
size = 32                & 12.46   & 20.43   & 16.77   & 13.92   & 22.96   & 18.21   & 17.20   & 28.27   & 22.12  \\
size = 64                & 12.54   & 20.59   & 16.61   & 14.04   & 23.17   & 18.34   & 17.42   & 28.63   & 22.55  \\
size = 128               & 12.48   & 20.48   & 16.72   & 13.93   & 23.00   & 18.13   & 17.20   & 28.36   & 22.17  \\
size = 256               & 12.85   & 21.12   & 17.12   & 14.27   & 23.56   & 18.81   & 17.72   & 29.02   & 23.39  \\ \bottomrule
\end{tabular}
\end{adjustbox}
\end{table*}

\vspace{5ex}
\begin{figure*}[!ht]
\begin{center}
\includegraphics[width=1\linewidth]{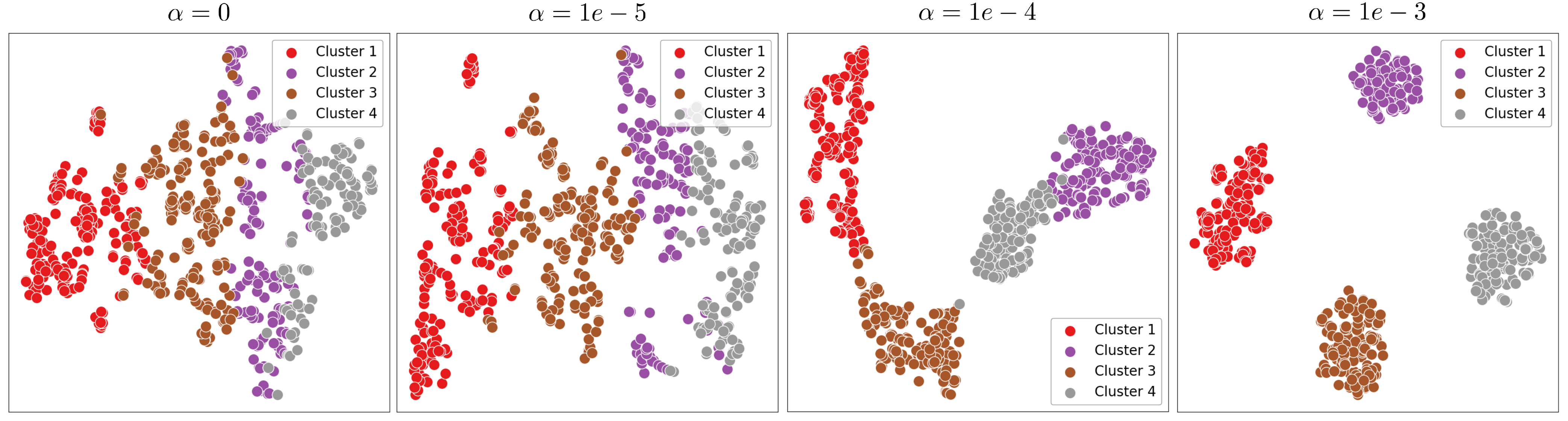}
\end{center}
\caption{t-SNE visualizations of latent representations \(\kappa_i = R_{pre; enc}(x^{1}_{i, w})\) with varying clustering loss weights \(\alpha\). Each subfigure represents the clustering outcome at a different value of \(\alpha\). From left to right, the subfigures correspond to \(\alpha = 0\), \(\alpha = 1e-5\), \(\alpha = 1e-4\), and \(\alpha = 1e-3\), respectively. 
}
\end{figure*}

\clearpage

\begin{figure*}[ht]
\begin{center}
\includegraphics[width=1\linewidth]{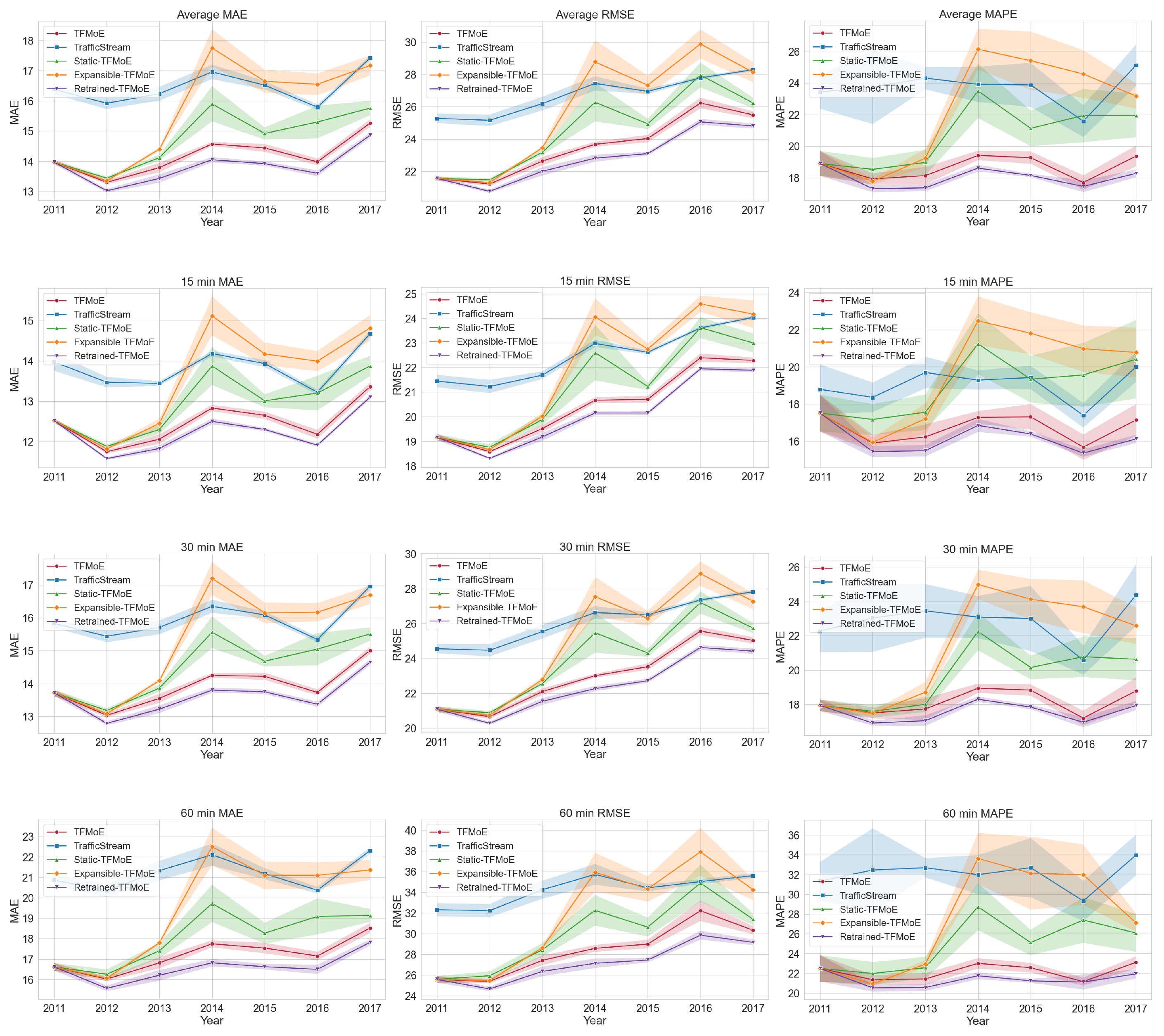}
\end{center}
\caption{The MAE, RMSE, and MAPE of traffic flow forecasting over consecutive years, with 1-sigma error bars.
}
\label{fig_all_time_grained}
\end{figure*}

%%%%%%%%%%%%%%%%%%%%%%%%%%%%%%%%%%%%%%%%%%%%%%%%%%%%%%%%%%%%

\end{document}